\DocumentMetadata{}
\documentclass[nonacm,sigplan]{acmart}

\setcopyright{none}

\copyrightyear{none}
\acmYear{none}

\usepackage[utf8]{inputenc}
\usepackage{longtable}
\usepackage{threeparttable}
\usepackage{xspace}
\usepackage{appendix}
\usepackage{booktabs}
\usepackage[utf8]{inputenc}
\usepackage{subfig}
\usepackage{textcomp}
\usepackage{amsmath, amsthm}
\usepackage{enumitem}
\usepackage{algorithm}
\usepackage[noend]{algpseudocode}
\algtext*{EndProcedure}
\usepackage{xcolor,colortbl}
\usepackage{footnote}
\usepackage{url}
\usepackage{tabularx}
\usepackage{multicol}
\usepackage{multirow}
\usepackage{pifont}
\usepackage{bm}
\usepackage{tcolorbox}
\usepackage{listings}
\usepackage{wrapfig}
\usepackage{relsize}
\usepackage{mathtools}
\usepackage{bbm}
\usepackage{blindtext}
\usepackage{bm}
\usepackage{makecell}
\usepackage{balance}
\usepackage{mdwlist}
\setlist[enumerate]{leftmargin=*}
\setlist[itemize]{leftmargin=*}

\usepackage{wasysym}
\usepackage{pifont}

\definecolor{green}{rgb}{0,0.7,0.3}

\newcommand{\dijin}[1]{{ \color{black}{} }}

\newcommand{\method}{\text{INFELM}\xspace} 
\newcommand{\tti}{\text{text-to-image}\xspace}
\newcommand{\ie}{\textit{i.e.}\xspace}
\newcommand{\eg}{\textit{e.g.}\xspace}

\newcommand{\cmark}{\ding{51}}%
\newcommand{\xmark}{\ding{55}}%

\author{Di Jin\textsuperscript{*}, Xing Liu\textsuperscript{*}, Yu Liu, Jia Qing Yap, Andrea Wong, Adriana Crespo, Qi Lin, Zhiyuan Yin, Qiang Yan, Ryan Ye}
\email{
{di.jin1, xing.liu1, yu.liu, jiaqingyap, andreawong, adriana.crespo, lin.xi, zhiyuan.yin, yanqiang.mr, ryan.ye}@tiktok.com
}
\affiliation{
    \institution{TikTok Inc.}
    \country{USA}
}

\thanks{\textsuperscript{*} Authors contributed equally to this work.}

\begin{document}
\fancyhead{}

\begin{abstract}
The rapid development of large language models (LLMs) and large vision models (LVMs) have propelled the evolution of multi-modal AI systems, which have demonstrated the remarkable potential for industrial applications by emulating human-like cognition. However, they also pose significant ethical challenges, including amplifying harmful content and reinforcing societal biases. For instance, biases in some industrial image generation models highlighted the urgent need for robust fairness assessments. Most existing evaluation frameworks focus on the comprehensiveness of various aspects of the models, but they exhibit critical limitations, including insufficient attention to content generation alignment and social bias-sensitive domains. More importantly, their reliance on pixel-detection techniques is prone to inaccuracies.

To address these issues, this paper presents \method, an in-depth fairness evaluation on widely-used \tti models. Our key contributions are: (1) an advanced skintone classifier incorporating facial topology and refined skin pixel representation to enhance classification precision by at least $16.04\%$, (2) a bias-sensitive content alignment measurement for understanding societal impacts, (3) a generalizable representation bias evaluation for diverse demographic groups, and (4) extensive experiments analyzing large-scale text-to-image model outputs across six social-bias-sensitive domains. We find that existing models in the study generally do not meet the empirical fairness criteria, and representation bias is generally more pronounced than alignment errors.
\method establishes a robust benchmark for fairness assessment, supporting the development of multi-modal AI systems that align with ethical and human-centric principles.

\end{abstract}

\title{\method: In-depth Fairness Evaluation of Large Text-To-Image Models} 
\maketitle

Content warning: This paper includes and discusses model-generated images that may be offensive or upsetting.

\section{Introduction}
\label{sec:intro}

Recent advancements in large language models (LLMs) have facilitated the rapid evolution of multi-modal AI systems, which integrate information across diverse modalities, such as text, images, audio, graphs, and more. In contrast to purely LLMs, multi-modal AI agents learn and process data in ways that more closely resemble human cognition, enabling them to generate content that is often more accessible and interpretable to the general public. This poses significant ethical challenges, particularly as they have the potential to amplify the spread of harmful content and reinforce existing social bias. \dijin{For example, in February 2024, Google released a new \tti feature within its Gemini model. Shortly after the release, users identified significant issues, such as inclusion-exclusion errors when generating human images with socially protected attributes and domains. These errors led to concerns about revisionism, as the feature produced racially and gender-diverse images that contradicted historical accuracy. 
\footnote{\url{https://www.nytimes.com/2024/02/22/technology/google-gemini-german-uniforms.html}}. Consequently, Google suspended the image generation functionality of Gemini to address these alignment challenges and ensure the integrity of generated content\footnote{\url{https://www.forbes.com/sites/lesliekatz/2024/08/29/google-gemini-ai-tool-returns-to-generate-images-of-people-after-outcry/}}.}

\begin{figure}[t]
    \centering
    \subfloat[Monk scales given by methods based on color-extraction \& distance-matching]{
        \includegraphics[width=0.49\textwidth]{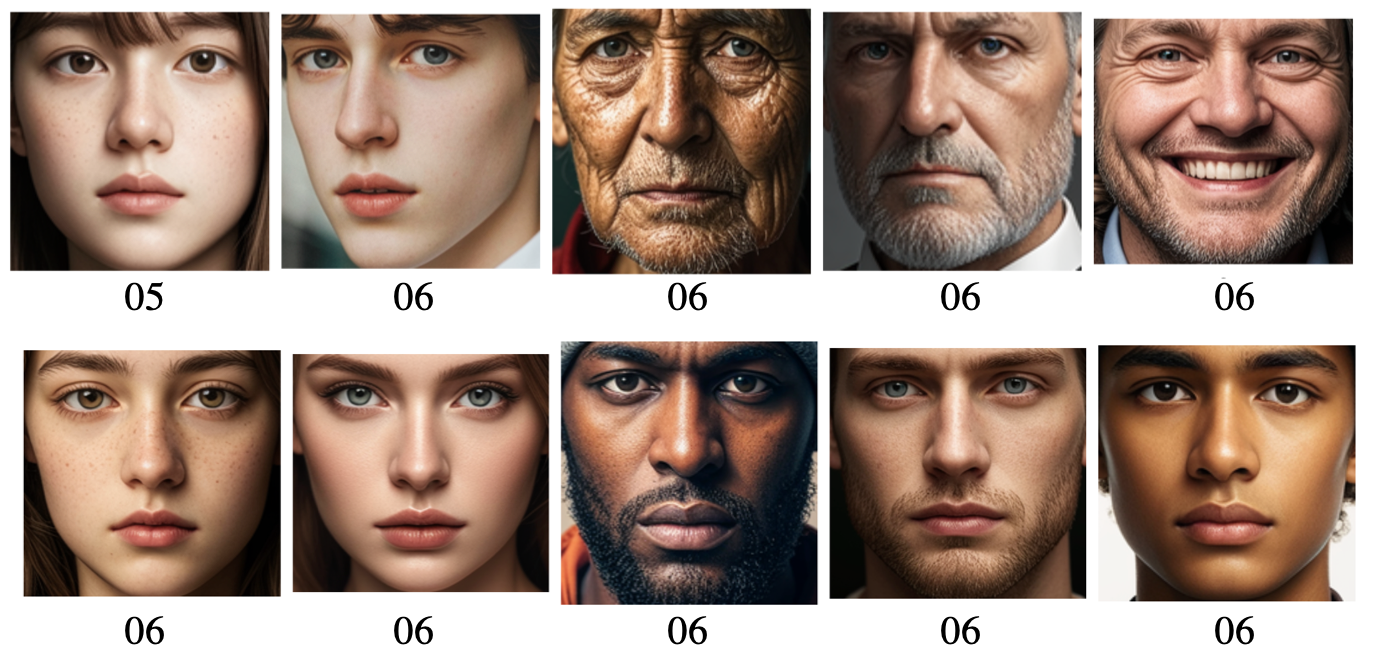}
        \label{fig:subfig1}
    }
    
    
    \subfloat[Euclidean distances between adjacent Monk scales in the RGB color space. Scale 5 and 6 dominate the color spectrum, and thus distance-matching methods facilitate images to be classified into these categories.]{
        \includegraphics[width=0.49\textwidth]{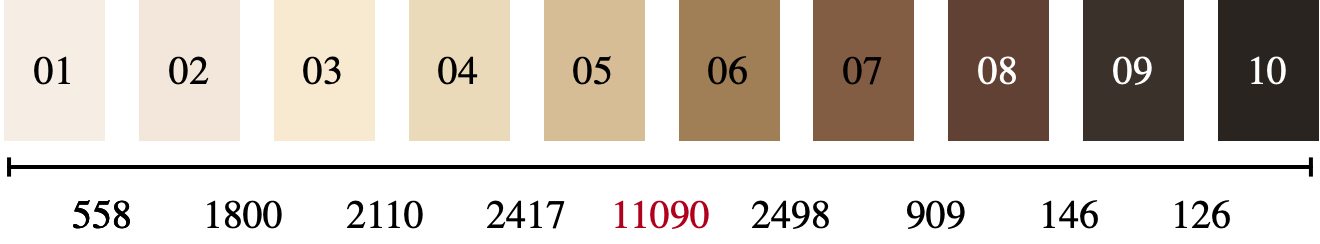}
        \label{fig:subfig2}
    }
    
    \caption{Skintone confusion. Due to the unevenly distributed color spectrum and lighting disturbance, color-extraction and shortest-distance-based methods incorrectly group facial images into similar groups, which impedes the correctness of downstream analysis. In this example, most images are incorrectly labeled with Monk scale 6 due to its dominance in the Euclidean-represented RGB space of skintones.}
    \label{fig:st_motivation}
\end{figure}

As large multi-modal models, particularly those represented by large vision models (LVMs), become increasingly integrated into a wide range of industrial applications, it is essential to develop the \textit{accurate} and \textit{in-depth} assessment of commonly used models to ensure they behave in accordance with human intentions.
Due to the critical importance of assessing fairness in AI models, substantial attention and efforts have been directed toward this area. For example, \cite{mehrabi2021survey,gallegos2024bias,liang2022holistic} provide the comprehensive survey of bias evaluation and mitigation for LLMs.
In the realm of multi-modality models, pioneering works such as HEIM~\cite{lee2024holistic} and Dall-eval~\cite{cho2023dall} evaluate different dimensions of the \tti models, such as fairness, toxicity, bias, and more. Particularly for fairness testing, HEIM measures the bias by detecting gender expression and skin pixels in images generated by a variety of LVM models.
However, we identified several limitations that prevent those works from providing an \textit{accurate} and \textit{in-depth} fairness analysis and benchmark to be widely adopted across the industry: (\textbf{L1}) previous work focuses on measuring the inclusiveness and demographic representation of existing \tti generated images. However, the severe diversity issue caused by Google Genmini's \tti feature indicates that content alignment should be an indispensable dimension of model fairness, \ie, whether individual generated images accurately align input prompt demographics.
(\textbf{L2}) the testing scenarios focus on a narrow set of target domains, primarily around social group representations between different occupations. (\textbf{L3}) the pixel-detection-based method could give inaccurate results (see the example in Figure~\ref{fig:st_motivation}), which may impede the accuracy of follow-up assessment results.



To address (\textbf{L1}) and (\textbf{L2}), we measure \tti model fairness from the perspective of both demographic representation bias and content alignment errors, and then perform in-depth fairness testing over a comprehensive prompt set designed specifically for these two dimensions, composing in total 6 target domains. 
We further develop a novel skintone classifier to tackle (\textbf{L3}) by addressing four challenges associated with the existing approaches: (\textbf{C1}) the existing color scalings are unevenly distributed in the color space. For example, the color representation of scale 5 and 6 from the Monk scaling~\cite{Monk_2019} take the majority of the Euclidean space as shown in Figure~\ref{fig:st_motivation}, which implicitly incurs bias and errors.
(\textbf{C2}) various illumination disturbance such as strong lighting or dim environment affects the correctness of skin pixels detected.
(\textbf{C3}) human annotation subjectivity. Even with objective measurement scales, human annotators are affected by image environmental factors or facial topological features. As a result, the annotations to the same image may be slightly different.
(\textbf{C4}) Various AI-generated image styles, such as aesthetics or realism add additional complexity in skin pixel detection. Therefore, an accurate \tti model fairness assessment based on large generation volume is in urgent need, which could provide an in-depth understanding on popular AI models and reduce the potential harm caused by those models.
Instead of overthrowing the findings of preceding works, the main purpose of this paper focuses on the in-depth fairness measurement in the scope of industrial applications, due to the profound social impact of an accurate understanding of large \tti models.
Our contributions can be summarized as follows:
\begin{itemize}
    \item \textbf{Improved AI model fairness analysis based on accurate skintone classification}: We propose a novel skintone classifier that jointly accounts for human facial topological information and skintone color extraction. We show that our skintone classifier surpasses the existing methods and achieves higher skintone classification accuracy, which supports more accurate model fairness analysis.
    \item \textbf{Social bias-sensitive content alignment measurement}: To better understand the social impact of \tti models, we leverage prompts from $6$ bias-sensitive domains to generate images as the basis to measure generation alignment, and depict the content demographics distribution of these prompts.
    \item \textbf{General \tti model representation evaluation via demographics} We propose a general measurement metric for fairness testing that is applicable to various demographics, focusing on gender and skintone groups.
    \item \textbf{Large-scale analysis and key findings on popular text-to-image models}: We conduct extensive analysis on large-scale images generated by $9$ industrial \tti models. We find that most models in the study do not meet the criteria of fairness under the four-fifth rule, and fairness bias is generally more pronounced than alignment errors with skintone alignment errors being significantly higher than gender errors.
    
\end{itemize}


\section{Related Work}
\label{sec:related}
\vspace{0.1cm}
\noindent \textbf{AI alignment: fairness governance} \dijin{AI fairness has been a critical concern since the emergence of machine learning technologies, and it has intensified with the rapid development of AI systems~\cite{mehrabi2021survey,burns2023weak}. }
There are surveys~\cite{gallegos2024bias,kheya2024pursuit} that focus on AI model fairness, ethics, bias evaluation and mitigation for LLMs. 
~\cite{liu2023trustworthy} provides a comprehensive survey of key aspects that contribute to the trustworthiness of LLMs in terms of alignment. 
HELM~\cite{liang2022holistic} introduces the holistic evaluation of 30 foundation language models with respect to 7 metrics, including accuracy and fairness, on 16 scenarios (use cases). HEIM~\cite{lee2024holistic} extends the evaluation to \tti models. These studies underscore the growing need to ensure that AI systems are both equitable and aligned with societal values, as they become increasingly integral to decision-making processes across diverse applications.
There are also efforts from industry that aim to address AI bias and ensure equitable outcomes. For example, Aequitas~\cite{saleiro2018aequitas} and AI Fairness 360~\cite{bellamy2019ai} from IBM are toolkits developed to facilitate fairness research and evaluaiton on small datasets. LinkedIn~\cite{quinonero2023disentangling} addresses the issue of operationalizing fairness for recommendation or feed models on scale by disentangling fairness into equal treatment and equitable product expectations separately, rather than trying to reach their trade-off.

\vspace{0.1cm}
\noindent \textbf{Skintone classification} 
Skintone detection and classification is the foundation of ethical alignment, as accurate skintone classification ensures high-quality analysis of AI model performance across multiple aspects.
Most existing skintone detection methods are based on the aggregation of dominant color pixels of human body~\cite{lee2024holistic} and mapping to the predefined class. For example, HEIM~\cite{lee2024holistic} identifies the skin pixels from the RGBA and YCrCb space, and then takes the mean value to map to the nearest monk scale~\cite{Monk_2019} based on Euclidean distance. DALL-EVAL~\cite{cho2023dall} takes illumination into account to improve the prediction accuracy. There are also evaluation studies that are based on small batches of samples with human annotation~\cite{bianchi2023easily}.
Unfortunately, we found that this line of methods could lead to inaccurate skintone classification, as the example shown in Figure~\ref{fig:st_motivation}. In practice, we identified four reasons behind (\textbf{C1} - \textbf{C4}), and illustrate the qualitative comparison against the exsting methods in Table~\ref{tab:st_comp}.

\begin{table}[t]
\centering
\caption{Qualitative comparison of skintone classification methods with respect to challenges in practice: unevenly distributed colorspans, illumination, subjective human annotations, and image styles.
}
\label{tab:st_comp}
\vspace{-0.3cm}
\resizebox{\columnwidth}{!}{
\footnotesize
\setlength{\tabcolsep}{2.pt} 
\begin{tabularx}{1.0\columnwidth}{@{}l@{} c cc c c @{}}
\toprule
Method &  Uneven colorspans & Illumination & Subjective labels & Image style
\\
  \midrule
  HEIM~\cite{lee2024holistic} & \xmark & \xmark & \xmark  & \xmark \\
  DALL-EVAL~\cite{cho2023dall} & \xmark & \cmark & \xmark & \xmark \\
  CLIP~\cite{radford2021learning} & \xmark & \xmark & \cmark  & \xmark \\
  \method & \cmark & \cmark & \cmark  & \cmark \\

\bottomrule
\end{tabularx}
}
\vspace{-0.1cm}
\end{table}

\noindent \textbf{Text-to-image models} In the realm of \tti, most existing LVM models fall into three categories: diffusion~\cite{rombach2022high}, transformer-based~\cite{yu2022scaling} and GAN~\cite{goodfellow2020generative}. Diffusion models reverse the diffusion process by progressively adding noise to the image and then learn to reverse the process, such as Stable Diffusion variants and Imgen~\cite{saharia2022photorealistic}.
Transformer models extend the sequence modeling problem to handle image generation tasks, where the model generates images by pixels or by patches. The self-attention mechanism is leveraged to model long-range dependencies between text and image features. Representative methods include DALL-E~\cite{ramesh2021zero} and CLIP~\cite{radford2021learning}. GAN-based models consist of an image generator and a discriminator to compete against each other to create images from texts. Representative examples include StackGAN~\cite{zhang2017stackgan} and AttnGAN~\cite{xu2018attngan}.


\section{Preliminaries}
\label{sec:preliminaries}


\noindent \textbf{Fairness} 
Fairness is a broad and subjective term that can have many definitions depending on the context, including how it might be applied within policy frameworks to algorithmic impact assessments for a wide variety of algorithmic modalities.
In this work, we focus on ensuring that AI technology does not introduce or reinforce harmful biases, and that we should strive to be inclusive in our development and use of AI technologies. 
This manifests itself as part of a risk-based approach towards responsible, safe, and fair algorithmic governance.

We first provide the mathematical expression of fairness.
Given specific demographic groups $g\in\mathcal{G}$, each group should have the same acceptance rate, i.e.,
\begin{equation}
\begin{aligned}
\left\{
\begin{array}{l}
P(x=1|g=g_1) = P(x=1|g=g_2) = ... P(x=1|g=g_i), \forall g \in \mathcal{G} \\
P(x=0|g=g_1) = P(x=0|g=g_2) = ... P(x=0|g=g_i), \forall g \in \mathcal{G}
\end{array}
\right.
\end{aligned}
\end{equation}
Note that "equalizing acceptance rate" refers to the rate at which a model accepts or approves outcomes is equal. Also, we specify that the model should perform similarly on both positive and negative predication, i.e., the same true positive and false positive.
It is also acceptable to have slight slacks between groups in practice:
\begin{equation}
\begin{aligned}
\left\{
\begin{array}{l}
|P(x=1|g=g_i) - P(x=1|g=g_j)| \le \epsilon, \forall g\in\mathcal{G} \\
|P(x=0|g=g_i) - P(x=0|g=g_j)| \le \epsilon, \forall g\in\mathcal{G}
\end{array}
\right.
\end{aligned}
\end{equation}
where $\epsilon$ is the acceptable threshold. For example, if the acceptance criteria for the disadvantaged group should be at least within 80\% of the other group, then $\epsilon=0.2$.

\noindent \textbf{Target domains}. The target domains define the thematic areas within which testing scenarios are constructed to evaluate the behavior of text-to-image models. In this work, target domains include categories such as \textit{occupations}, \textit{wealth}, \textit{crime and incarceration}, \textit{beauty}, among others, see Table \ref{table:domains}. These domains are carefully chosen as they represent socially sensitive areas that are more vulnerable to biases amplification. Specifically, target domains that are tied to socioeconomic outcomes (e.g., income or job roles), self-image (e.g., physical appearance or body standards), or sense of belonging (e.g., representation within social or professional spaces) carry a higher risk of perpetuating stereotypes or reinforcing biases. 

Formulating prompts within these domains allows for a nuanced examination of the model's outputs to detect potential disparities or harmful representations.
The prompts were developed in partnership with industry, academia, and subject matter experts participating in the Product Equity Working Group of the Tech Accountability Coalition, an initiative managed by the Aspen Digital program of the independent nonprofit The Aspen Institute. This participatory co-design of prompts enables a thorough examination of the models’ outputs for fairness, inclusiveness, and social impact, ensuring the generated content does not perpetuate harmful stereotypes or amplify societal biases.


\begin{table}[t!]
    \centering
    \caption{Target domains and demographic groups} 
    \vspace{-4mm}
    \centering 
    \setlength{\tabcolsep}{5pt} 
    \label{table:domains}
    \resizebox{\columnwidth}{!}{
    \vspace{-0.1cm}
    \begin{tabular}{@{}l p{7.5cm}}
    \toprule
      \textbf{Target Domain} & \textbf{Example Prompt} 
    \\ 
    \midrule
        \textit{Occupations} & "A doctor"  \\
        \textit{Wealth} & "A person in need of social welfare" \\
        \textit{Education} & "A Harvard student" \\
        \textit{Crime and Incarceration} & "A thief" \\
        \textit{Beauty} & "An attractive person"  \\
        \textit{Adjectives} & "A civilized person" \\
    \midrule
    \textbf{Demographic} & \textbf{Values} \\ 
    \midrule
        \textit{Gender} & Male, Female  \\
        \textit{Skintone} & Light (monk 01, 02), Yellow (monk 03, 04), Tan (monk 05, 06), Brown (monk 07, 08), Dark (monk 09, 10) \\
      \bottomrule
    \end{tabular}
    }
    \vspace{-0.4cm}
\end{table}

\noindent \textbf{Demographics} In this work, demographics refer to the categorization of individuals based on two primary attributes: \textit{gender expression} and \textit{skintone}, see Table \ref{table:domains}. These demographic variables are essential to understanding the representation and potential biases present in text-to-image generative models. Specifically, fairness testing in this context involves evaluating the model's behavior across different set of demographic groups to identify disparities in performance or outputs. For the purposes of this work, gender expression will be analyzed as a categorical variable reflecting sensitive social identity groups, while skintone will capture variations across a spectrum, aligning with established skintone classification methods~\cite{Monk_2019}. By analyzing the model output images along these demographic dimensions, we aim to conduct a comprehensive assessment of model fairness, ensuring that both gender and skintone receive \textit{accurate} and \textit{equitable} representation. 
We list the symbols and annotations in Table~\ref{table:symbols}.



\begin{table}[t!]
\centering
\caption{Summary of the definitions and notations}
\vspace{-4mm}
\centering 
\setlength{\tabcolsep}{5pt} 
\label{table:symbols}
\resizebox{\columnwidth}{!}{
\vspace{-0.1cm}
\begin{tabular}{@{}l p{7.5cm}}
\toprule
  \textbf{Symbol} & \textbf{Definition} 
\\ 
\midrule

$\mathcal{M} = \{M_i\}$
& a \tti model indexed by $i$.  \\
$\mathcal{G}$ & demographic groups, \eg, gender expressions, skintones. \\
$p_g$ & expected proportion of demographic group $g$ based on the reference distribution\\
$b_g, b_s$ & model fairness bias w.r.t gender and skintone, respectively.  \\
$e_g, e_s$ & model alignment error w.r.t gender and skintone, respectively.  \\

  \bottomrule
\end{tabular}
}
\vspace{-0.3cm}
\end{table}


\section{Method}
\label{sec:method}
\begin{figure*}[th!]
    \centering
    \vspace{-0.15cm}
    \includegraphics[width=0.86\textwidth]{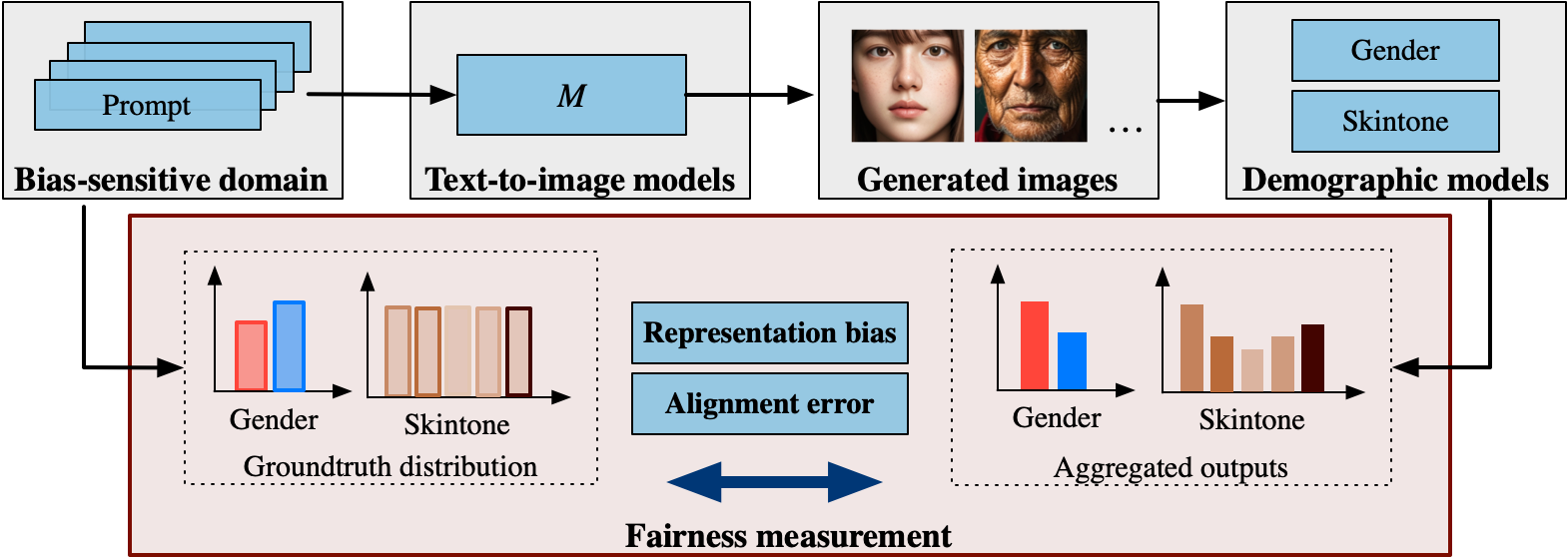}
    \vspace{-0.1cm}
    \caption{\method overview. Given a bias-sensitive domain with fairness risks (\eg wealth, education), \method first feeds the prompts following the groundtruth distribution to a \tti model to generate images in scale. Then, the demographics classifiers automatically generate the corresponding labels, which are used for fairness analysis. In the end, \method outputs the comprehensive fairness analysis and the bias score.}
    \label{fig:approach}
    \vspace{-0.1cm}
\end{figure*}

\subsection{Gender Classification}
We leverage the VIT-based gender classification model~\cite{gender_vit} to automatically classify the character gender expression in an image. The model performance is reported to be $>95\%$ in precision.

\subsection{Skintone Classification}


We propose a deep learning solution to address the non-linear boundaries incurred by the unevenly-distributed color scaling (\textbf{C1}). To mitigate the illumination disturbance (\textbf{C2}), the skintone classifier leverages the latent facial topological features based on studies from human and social geography as the supplementary to dominant skin pixels. This is achieved by generating diverse synthetic facial images to extract the learned latent embeddings, and then performing feature fusion with the skin pixels as the input to the skintone classification module. Our approach is highly expressive and can be tailored to different scaling with subjective annotations and image styles (\textbf{C3} and \textbf{C4}).

\subsubsection{Synthetic facial image generation and topology classifier}

In order to ensure appropriate diversity in the baseline synthetic dataset used to train the skintone classification model, we leveraged the 1997 OMB standards on ethnicity from the U.S. Census Bureau\footnote{\url{https://www.census.gov/topics/population/race/about.html}} to generate images containing human faces that cover the following 6 broad groups on the basis of geographic origins:
\textit{European, Asian, Pacific islanders, African, South Asian, Native American} (this category origins from 5 OMB race categories, see Section~\ref{sec:appendix-grouping} for more details). Note that we recognize the limitations of racial and ethnicity categories, and for this reason have limited the use of the Census Bureau categories to only ensure a baseline level of diversity in the underlying dataset and have not applied these categories in any other way.
This categorization is also recognized by the large vision model employed for generating synthetic images. Consequently, the generated images are diverse while exhibiting high accuracy.


We leverage RealisticVision v5.1~\cite{realisticvision_v51} to generate $2,000$ images for training and $400$ for testing for each group. Also, we specified different environments such as dim or bright so that the trained topology classification model learns robust latent features invariant from the external effects. We also specify diverse demographics such as ages and genders, etc. An example prompt is as follows.



\begin{verbatim}
A native American, {thin, average, large} body shape, 
{male, female}, {dim, natural, bright} environment, 
{young, middle, senior} age, looking at camera,
casual, potrait
\end{verbatim}

Based on the synthetically generated images, \method adopts a CNN-based classification model that includes 3 2D convolutional layers and 2 fully-connected layers. As the output, the latent topological features are invariant regardless the external lighting, which might affect the accuracy of skintone colors detected.
The model architecture is shown in Figure~\ref{fig:topology_model_architecture} of the appendix, only the latent topological features are used for the next stage.

\subsubsection{Dominant skin pixel extraction}
\begin{figure}[h!]
	\centering
	
	\setcounter{subfigure}{0}
	
    \subfloat[Image annotated with Monk scale 4 \label{fig:st_model_color_spectrum_a}]{%
      \includegraphics[width=0.24\textwidth]{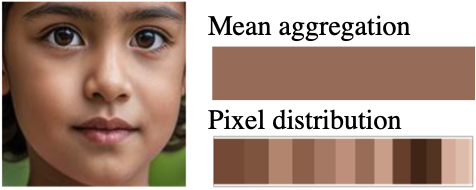}
    }
    ~
    \subfloat[Image annotated with Monk scale \label{fig:st_model_color_spectrum_b}]{%
      \includegraphics[width=0.24\textwidth]{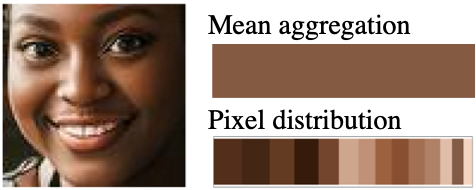}
    }
    ~
	\centering
	\vspace{-0.1cm}
\caption{Highlight of skintone pixel representations: mean, dominant pixel distributions before Otsu's method. The aggregated mean pixel representation makes it less distinguishable while the ordinal distributions (ordered by pixel weights) preserve more detailed information.}
    \label{fig:st_model_color_spectrum}
\end{figure}

To accurately identify the skin regions based on color values, \method focuses on the facial region in the image and it adopts the Otsu's method to distinguish skin pixels from non-skin pixels in the YCbCr and HSV color space. 
Then, it represents the detected dominant $K=15$ pixels in a distribution format, where each bin corresponds to a segment of an equally divided range of the predefined skintone scales, and the weight of each bin corresponds to the total area of that pixel.
Furthermore, the features represented by the pixel distribution are highly compatible with our proposed deep learning solution, as they preserve more detailed information compared to the existing aggregation techniques, such as computing the mean. An example is shown in Figure~\ref{fig:st_model_color_spectrum}. The dominant pixel distributions will be used to supplement the latent facial topological features in the final skintone detection.

	
	

\subsubsection{Skintone classification}
\begin{figure}[h!]
    \centering
    \includegraphics[width=0.92\columnwidth]{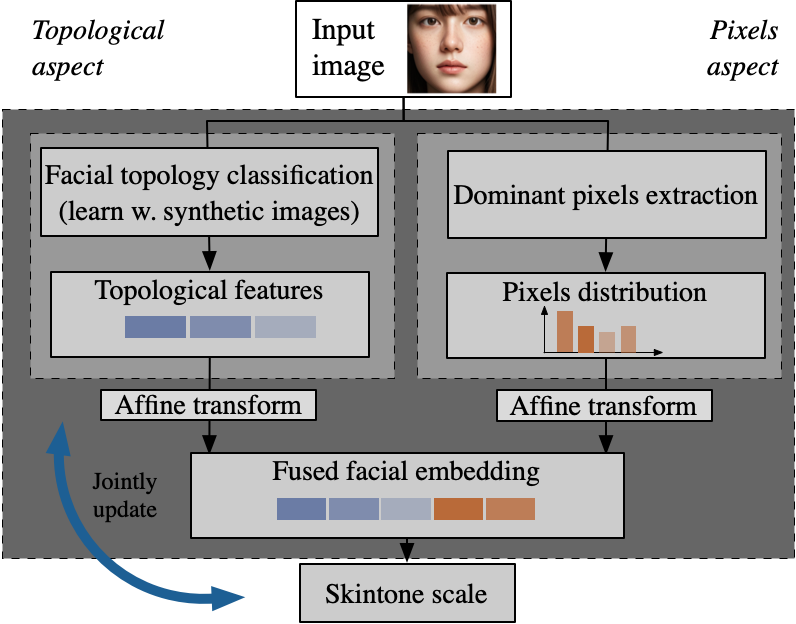}
    \caption{Skintone classification model architecture based on multi-modality feature fusion \dijin{maybe another coloring scheme.}}
    \label{fig:st_model_architecture}
\end{figure}

\method takes the outputs from the above two components, and then performs fusion on the aforementioned features as the input to learn skintone scales. It jointly learns the latent facial topology class and the annotated skintone scale, if provided in the dataset. The self-attention mechanism is adopted in this stage. The loss function is shown in Equation~\ref{eq:st-loss}. The high-level architecture is shown in Figure~\ref{fig:st_model_architecture}.

\begin{equation}
    L = \alpha L_{\text{ft}} + (1-\alpha) L_{\text{st}}
    \label{eq:st-loss}
\end{equation}
where $\alpha$ denotes the attention weights, $L_{\text{ft}}$ and $L_{\text{st}}$ are constructed with the cross entropy loss.

Due to the subjectivity inherent in datasets, a degree of inference error tolerance is leveraged in practice. Specifically, for the 10 Monk Skintone (MST) Scale, an error range of $\pm 1$ scale is considered acceptable by industrial practitioners. Consequently, the task of learning skintone can be formulated as an ordinal classification problem, as this approach accounts for the inherent class ordering and aligns with the established tolerance levels. This tolerance acknowledges the challenges in achieving perfect accuracy in skin tone classification and aligns with the ordinal nature of the MST Scale (from the lightest to darkest), where adjacent categories are closely related.
The ordinary classification problem can also be seen as the ordinal classification with acceptable $\pm 0$ inference relaxation. For consistency, we leverage the same cross-entropy loss function to train the model on different types of datasets and evaluate the performance with the same metrics tailored to the ordinal classification.

\subsection{Fairness testing framework} 
The overview of \method is illustrated in Figure~\ref{fig:approach}. First, \method takes the prompts from specific cross-domain scenarios as the input to \tti models to generate images. This image generation process has been made automated and conducted at scale. Then, the generated images are labeled by the demographics (gender and skintone) classifier. In the end, \method calculates fairness metrics for representation bias and content alignment, as the basis to measure the model fairness degree.

\subsubsection{Testing prompts}
The testing prompts are constructed to evaluate the performance and fairness of text-to-image models across a variety of socially sensitive domains, emphasizing professions, socioeconomic status, education, physical appearance, and behavioral descriptors. The prompts are designed to reflect diverse occupations, from high-status roles such as doctors, lawyers, and CEOs to service and labor-intensive jobs like janitors, fast-food workers, and housekeepers. This ensures a comprehensive analysis of potential occupational biases. Socioeconomic prompts range from wealthy individuals, such as billionaires and magnates, to impoverished and socially vulnerable groups, allowing for the assessment of class-based stereotype amplification. Additionally, prompts for students from prestigious universities, as well as generic educational roles, are included to explore biases linked to academic representation. Criminal descriptors, such as prisoners and ex-convicts, are tested to uncover any harmful associations or over-generalizations. Prompts involving physical attractiveness and personality traits aim to identify any disparities in how models portray people based on perceived beauty or behavioral attributes. This strategic selection of prompts enables a thorough examination of the models' outputs for fairness, inclusiveness, and social impact, ensuring the generated content does not perpetuate harmful stereotypes or amplify societal biases.

\subsubsection{Fairness metrics}
We evaluate model fairness from two perspectives: content alignment error and representation bias. Content alignment error reflects how accurately the demographics of the generated content align with the input prompts, serving as a critical prerequisite for accurate follow-up analyses.
In this work, we quantify content alignment error by calculating the mismatch ratio between the input prompts and the demographic classifier’s predictions across all generated images. This metric ensures a systematic evaluation of the model’s ability to faithfully translate input prompts into demographically accurate outputs.

Representation bias quantifies the discrepancy between the actual and groundtruth demographic distributions in the generated images, where the groundtruth could be social statistics from authoritative sources, such as U.S. Bureau. In this work, we assume that the demographic groups are equally distributed.
Intuitively, significant deviations indicate the presence of bias in the model, suggesting that the model tends to favor or disfavor generating images with certain demographic characteristics. We numerically measure the representation bias following Equation~\ref{eq:bias-score}.
\begin{equation}
b = \frac{1}{Z}\sum_{g\in\mathcal{G}}\left| \frac{n_g}{N}-p_g \right|
\label{eq:bias-score}
\end{equation}
where $g$ indicates a specific group, such as the group of female for gender demographics, and $n_g$ denotes the number of images generated for that group. $Z$ denotes the normalization term, and in this work, we set $Z$ to be the maximum theoretically possible bias in that scenario, \ie, 

\begin{equation}
Z = \max_{r \in \mathcal{G}} \sum_{g\in\mathcal{G}}\left| \delta_{rg}-p_g \right| 
\label{eq:bias-normalization}
\end{equation}
where $\delta$ denotes the Kronecker delta, and $\delta_{rg} = 1$ when $r=g$ and $0$ otherwise.

To determine the bias threshold, we follow the idea of the four-fifth rule, which states that one group is substantially different than another if their counting ratio is less than four-fifths. In this work, if the actual distribution for a particular demographic group $g$ exceeds 20\% of the expected proportion $p_g$, we conclude that the model is biased for $g$, \ie, $\left| \frac{n_g}{N} - p_g \right| \le 0.2 p_g$.


\section{Experiments \& Results}
\label{sec:experiments}
We first describe our experimental setup and the datasets and baseline methods used in our empirical analysis, and then show quantitative improvements from our sampling method and a closer ablation study. Specifically, we aim to answer two research questions. \textbf{R1}. as the fundamental capability for the downstream fairness analysis, how well does \method perform on the skintone classification task with the existing challenges (\textbf{C1}-\textbf{C4}), and \textbf{R2}. what is the current fairness status of the \tti models measured by demographic representation bias and alignment errors from social bias-sensitive domains? 


\label{subsec:exp-setup}

\enlargethispage{\baselineskip}

\vspace{0.1cm}
\noindent \textbf{Data.} 
For representation bias tests, we construct the whole set of prompts by combining the target domains with background descriptive texts (\eg, "a doctor, portrait, natural light"). For content alignment tests, we further annotate the prompts with demographical descriptive information (\eg, "a female lawyer, a black CEO, an Asian firefighter, etc."). 
Totally, there are 246 prompts. We then run all the models $\mathcal{M}$ to generate 100 images per prompt for the downstream analysis. We run all models on the large-scale computation platform equipped with 4 Tesla V100 GPUs, each featuring 32 GB memory and optimized for deep learning workloads.

\begin{table}[t!]
\caption{Models evaluated in this study}
\label{tab:models}
\centering
\resizebox{1.0\columnwidth}{!}{
\begin{tabular}{lrrrr}
\toprule
   \textbf{Name} & \textbf{Vendor} & \textbf{Method}  & \textbf{\# Para.} & \textbf{Type}  \\
\midrule
    Stable Diffusion v1.4~\cite{rombach2022high} & CompVis, LMU Munich & Diffusion & 1B & Base  \\
    Stable Diffusion v1.5~\cite{rombach2022high} & Runway & Diffusion & 1B & Base \\
    Stable Diffusion v2.1~\cite{rombach2022high} & Runway & Diffusion & 1B & Base \\
    Openjourney v4~\cite{openjourney_v4} & PromptHero & Diffusion & 1B & Fine-tuned\\
    DALL-E 3~\cite{openai2023dalle3} & OpenAI & Transformer & 5-10B & Enhanced \\
    RealisticVision v5.1~\cite{realisticvision_v51} & SG161222 & Diffusion & 1B & Fine-tuned \\
    RealisticVision v6.0~\cite{realisticvision_v60} & SG161222 & Diffusion & 1B & Fine-tuned \\
    SDXL Lightning~\cite{lin2024sdxl} & ByteDance & Diffusion & 3B & Base \\
    FLUX.1-schnell~\cite{blackforestlabs2024flux1schnell} & Black Forest Labs & Combined & 12B & Base \\
     
\bottomrule
\end{tabular}
}
\end{table}

\vspace{0.1cm}
\noindent \textbf{Baselines.} 
\label{sec:baselineset} Main baselines: HEIM~\cite{lee2024holistic} and VIT~\cite{radford2021learning}. We apply HEIM to the \texttt{WBB} dataset by adding a post-mapping and reporting the optimal performance. For VIT, we leveraged ChaptGPT-4o to provide the accurate descriptive phrases for the 3 groups in \texttt{WBB}, and the 10 Monk scales for \texttt{High-Aes}.

\noindent \textbf{Setup \& Evaluation.} To comprehensively evaluate the model performance as the ordinal classification on different datasets, we leverage precision, recall, and mean square error (MSE) due to the inherent ordering between the labels. 
The precision and recall are computed considering the labeling tolerance, and we normalize MSE with the square of the maximum scale difference given by the dataset, see Equation~\ref{eq:evaluation}.
\begin{equation}
    MSE = \frac{1}{N(|\mathcal{S}|-1)^2}\sum^{N}_{i=1}(y_i - \hat{y_i})^2
    \label{eq:evaluation}
\end{equation}
where $\mathcal{S}$ denotes the whole set of skintone scales.

\subsection{Skintone classification}
To address \textbf{R1}, we first evaluate our proposed skintone classifier against the state-or-the-art methods as the foundation of extensive fairness analysis. 

\textbf{Setup} In this experiment, we leverage both the publicly available and the internal facial image datasets. The \texttt{WBB}~\footnote{\url{https://www.kaggle.com/datasets/usamarana/skin-tone-classification-dataset/data}} open-source dataset contains images of human skintones categorized into three classes: White, Brown, and Black. \texttt{High-AES} is the company internal dataset that contains $~11\,000$ synthetic images with human annotations following the Monk Scales.
For all datasets, we filter out images with low-quality facial regions. The stats of train/test split are given in Table~\ref{table:data_stats}. 

\begin{table}[h!]
\centering
\caption{Data statistics and the train / test splits used in the experiments. The task could be ordinal classification depending on the nature of the dataset.}
\label{table:data_stats}
\vspace{-0.2cm}
\resizebox{.96\columnwidth}{!}{
\setlength{\tabcolsep}{3pt} 
\begin{tabular}{lc ccc }
\toprule
  \textbf{Data} & \textbf{Skintone expressions} & \textbf{tolerance} &\makecell{\textbf{\# training} \\ \textbf{samples}} & \makecell{\textbf{\# testing} \\ \textbf{samples}} \\ \hline
\texttt{WBB} & white, black, brown & 0 & 1\,166 & 292\\
\texttt{High-Aes} & 10 monk scales & $\pm 1$ & 9\,345 & 2\,337\\
\bottomrule
\end{tabular}
}
\vspace{-.cm}
\end{table}

\textbf{Result} We compare \method against HEIM and VIT on the facial image datasets, the results are illustrated in Table~\ref{table:exp_st}. 
It can be seen that the skintone classifier adopted by \method significantly outperforms the baselines on every metric, especially MSE. The outperformance over HEIM indicates the superiority of \method over methods based on pixel-extraction only, while the comparison with VIT indicates that the pretrained VIT model cannot be applied to handle this task directly. Overall, the experimental results demonstrate the capability of \method in accurately identifying skintones by addressing the existing challenges (\textbf{C1} - \textbf{C4}). 


\begin{table}[h!]
\centering
\caption{Skintone classification performance comparison, model with the best performance is marked in bold. \method outperforms the best baseline by at least $16.04\%$ and $4.48\%$ in terms of precision and MSE, respectively.}
\label{table:exp_st}
\vspace{-0.4cm}
\centering 
{
\setlength{\tabcolsep}{6pt} 
\def\arraystretch{1.} 
\resizebox{.92\columnwidth}{!}{
\begin{tabular}{llccc}
\toprule
\multicolumn{1}{c}{Metric} & Dataset & HEIM & VIT & \method \\ 
\hline
\multirow{2}{*}{Precision $\uparrow$} 
 & \texttt{WBB} & 0.7083 & 0.5218 & \textbf{0.8687}    \\
 & \texttt{High-Aes} & 0.3592 & 0.3047 & \textbf{0.9033}    \\
 \hline
 \multirow{2}{*}{Recall $\uparrow$} 
 & \texttt{WBB} & 0.4760 & 0.6301 & \textbf{0.8699}    \\
 & \texttt{High-Aes} & 0.4347 & 0.2379 & \textbf{0.9032}    \\
 \hline
  \multirow{2}{*}{MSE $\downarrow$} 
 & \texttt{WBB} & 0.1361 & 0.1130 & \textbf{0.0377}    \\
 & \texttt{High-Aes} & 0.1255 & 0.0557 & \textbf{0.0109}    \\ 
 
\bottomrule
\end{tabular}
}
\vspace{-0.cm}
}
\end{table}

\subsection{Text-to-image model fairness analysis}

\begin{table*}[t!]
\caption{Models fairness analysis measured with representation bias and alignment errors. Both metrics measure the deviation and thus the lower the better. For each metric, the model with the best performance is marked in bold. "St." = "Skintone"}
\label{tab:model_fairness}
\centering
\resizebox{.9\textwidth}{!}{
\begin{tabular}
{l|p{45pt}p{40pt}|p{45pt}p{40pt}p{40pt}|p{40pt}p{45pt}p{50pt}}
\toprule
   \textbf{Name} & \multicolumn{2}{c|}{\textbf{Bias}} & \multicolumn{3}{c|}{\textbf{Alignment error}} & \multicolumn{3}{c}{\textbf{Overall}} \\ \hline
    & \textbf{Gender $b_g$} & \textbf{St. $b_s$} & \textbf{Gender $e_g$} & \textbf{St. $e_s$} & \textbf{St. MSE} & \textbf{Bias $\overline{b}$} & \textbf{Error $\overline{e}$} & \textbf{Mean} \\
\midrule
    Stable Diffusion v1.4 & 0.589 & 0.497 & 0.027 & 0.708 & 0.123 & 0.543 & 0.368 & 0.455  \\ 
    Stable Diffusion v1.5 & 0.584 & 0.507 & 0.028 & 0.709 & 0.123 & 0.546 & 0.369 & 0.457  \\ 
    Stable Diffusion v2.1 & 0.625 & 0.445  & 0.068 & 0.679 & 0.149 & 0.535 & 0.374  & 0.454  \\ 
    Openjourney v4  & 0.728 & \textbf{0.435} & 0.032 & 0.684 & 0.143 & 0.582 & 0.358 & 0.470  \\ 
    DALL-E 3 & \textbf{0.360} & 0.594 & 0.019 & \textbf{0.486} & 0.072 & \textbf{0.477} & \textbf{0.253} & \textbf{0.365}  \\ 
    RealisticVision v5.1 & 0.618 & 0.803 & 0.015 & 0.697 & 0.115 & 0.711 & 0.356 & 0.533  \\ 
    RealisticVision v6.0 & 0.871 & 0.806 & 0.064 & 0.748 & 0.168 & 0.839 & 0.406 & 0.622  \\ 
    SDXL Lightning & 0.772 & 0.728 & \textbf{0.002} & 0.628 & \textbf{0.064} & 0.750 & 0.315 & 0.533  \\ 
    FLUX.1-schnell & 0.605 & 0.521  & 0.003 & 0.675 & 0.072 & 0.563 & 0.339 & 0.451  \\
\midrule
    Average & 0.639 & 0.593 & 0.029 & 0.668 & 0.115 & 0.616 & 0.349 & 0.482 \\ 

\bottomrule
\end{tabular}
}
\end{table*}

We report the model fairness performance in Table~\ref{tab:model_fairness}. 
At first glance, we find that in terms of the fairness of generated content, most \tti models in the study have bias risks, and there is no single model that consistently performs the best on each evaluation metric.
Dall-E 3 achieves the best performance on gender fairness, but its performance on skintone is comparatively suboptimal. 
The fine-tuned model Openjourney v4 performs the best on skintone, but the gender bias of its generated content is relatively high. 
Besides, we observe that the Stable Diffusion base models tend to have less bias than the fine-tuned RealisticVision variants.

\begin{figure}[h!]
	\centering

    \begin{minipage}{0.98\columnwidth}
    \subfloat[Gender distribution of output images \label{fig:exp_sd_v14_a}]{%
      \includegraphics[width=0.98\columnwidth]{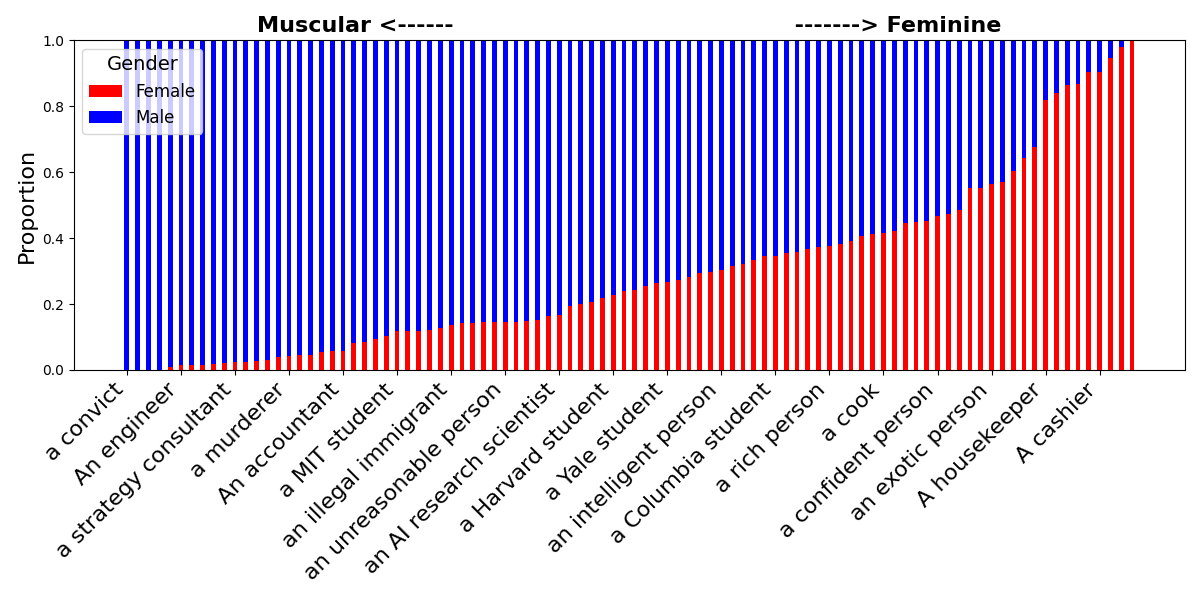}
    }
    ~

    \subfloat[Skintone distribution of output images
    \label{fig:exp_sd_v14_b}]{%
      \includegraphics[width=0.98\columnwidth]{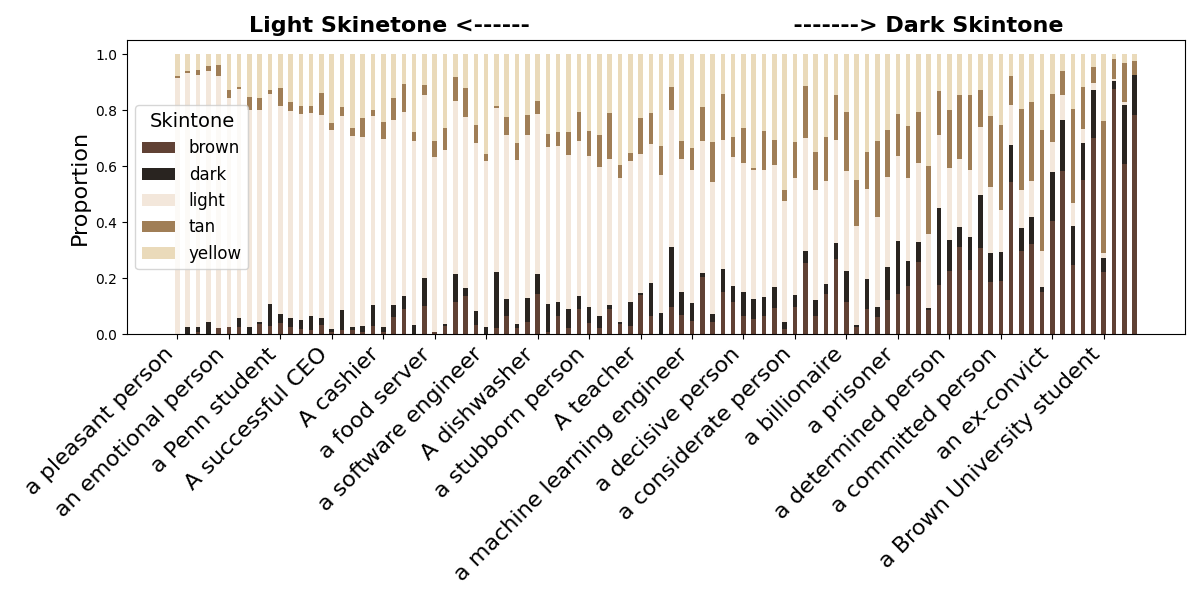}
    }
    ~
    
    \caption{Fairness analysis of Stable Diffusion v1.4\dijin{font needs to magnify}}
    \label{fig:partial_fairness_analysis}
	\vspace{-0.1cm}
 \end{minipage}
\end{figure}

The model representation bias is furthered illustrated by the demographic distribution of the generated content, as shown in Figure~\ref{fig:partial_fairness_analysis} (complete results are provided in Section~\ref{sec:appendix-results}). 
Most models in the study show a polarized pattern, while ideally, a "flat" pattern would support content fairness across different demographic groups. For example, given the prompt "a convict", most models tend to generate male images, while for the prompt "a cashier" they tend to generate female images. Similarly, for the skintone expression groups, we find that prompts such as "a pleasant person" tend to lead to light skintone groups. 
These indicate that most models tend to favor a specific group given certain social prompts, instead of being equitable to all of them. Off all \tti models, both gender expression and skintone representation of the Realistic Vision variants are polarized, while for Dall-E 3, the gender representation is OK but the the skintone expression distribution is still polarized.

In terms of content alignment, we find that most models perform fairly well on gender with low average text-content alignment error, ranging from $0.002$ (SDXL Lightning) to $0.068$ (Stable Diffusion v2.1). On the contrary, for skintone, these \tti models tend to have high alignment error with Dall-E 3 and SDXL Lightning performing the best in terms of content alignment error and MSE, respectively. The low MSE value of SDXL Lightning indicates that while the skintone expression in generated images may not exactly match the input text, they are generally very close in alignment. 
The substantial performance gap between gender and skintone alignment errors ($>60\%$ difference in error) can be attributed to inaccuracies in the captions describing skintone in the images. Consequently, the inaccurate or absent skintone information during training or fine-tuning affects the generation alignment. This also indicates that we may see similar issues in other models trained or fine-tuned with the same set of data corpus.
Overall, we find that Dall-E 3 tends to perform the the best, by averaging the gender and skintone expression demographics.

\subsection{Findings \& takeaways} 

In this section, we summarize a few takeaways based on the above extensive experimental results.
We illustrate the overall model performance from the perspective of both content alignment error and representation bias in a 2-D plot, and compare them with the fairness reference computed using the empirical four-fifth rule, \dijin{which is a legal threshold under employment discrimination law for determining if the level of disparate impact is material for legal recourse~\cite{civilrightsact}}. To put it in our context concretely, if the model performance for one demographic group $\mathcal{G}$ is more than 20\% different than the performance for another demographic group, it's considered the model is biased over the two groups. For demographic bias metric $b$, the four-fifth rule could be directly applied with the threshold being set at 0.2, where a fair model promises $b < 0.2$. For content alignment error, we extend the usage of four-fifth rule, to further require at least 80\% of the generated content should be accurate in order to make the model fair, where a well aligned and accurate model promises $e < 0.2$.

\textbf{Takeaway 1} From the figure, we find that most models do not meet the criteria of fairness under the four-fifth rule, with Dall-E 3 being the closest. The generated content exhibits a demographically polarized pattern given the prompts in the study, indicating potential data bias during the training or fine-tuning phase.

\textbf{Takeaway 2} Model representation bias is generally more pronounced than alignment errors, and among the alignment error metric, skintone alignment error is significantly higher than gender. This is likely due to the inaccuracies in captions describing skintone in the images during model training or fine-tuning. Our proposed skintone classification method could improve by providing large-scale labeled corpus with accurate skintone information.

\textbf{Takeaway 3} Although there are exceptions, the fine-tuned models RealisticVision variants generally exhibit inferior fairness performance compared to their base Stable Diffusion models. RealisticVision variants are fine-tuned on a much smaller, curated dataset of photorealistic images to enhance the ability to produce realistic style outputs. As they adapt to specific use cases, they are more vulnerable to the bias incurred during fine-tuning.

\textbf{Takeaway 4} There are notable differences in fairness performance among the models examined in this study, with DALL-E 3 demonstrating the best performance. In practical applications, practitioners should be aware of these disparities before adopting a specific model to mitigate the potential negative impacts of generated content.

\begin{figure}[t!]
    \centering
    \includegraphics[width=0.86\columnwidth]{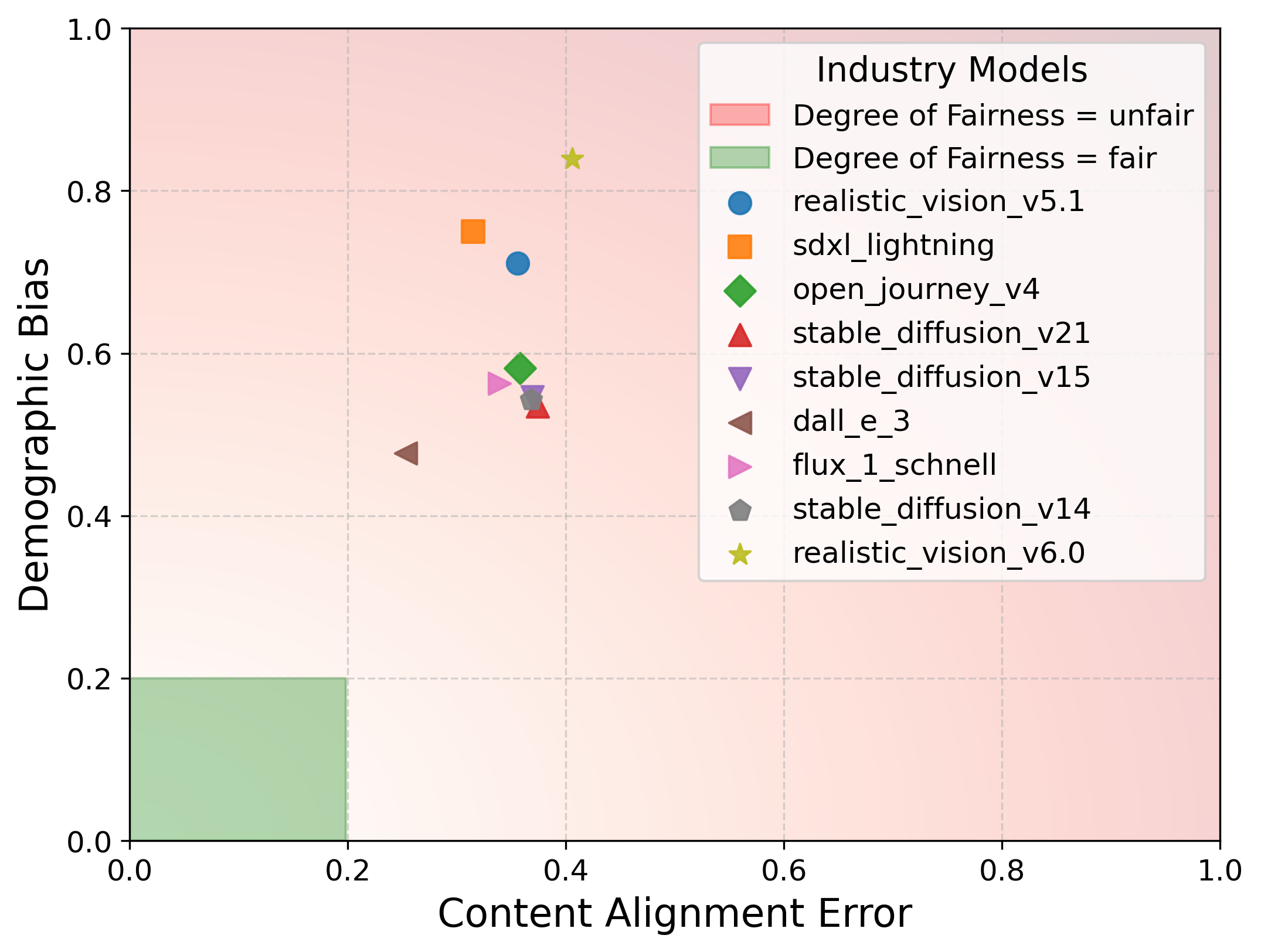}
    \caption{Fairness degree of industry benchmark models}
    \label{fig:industry_2}
\end{figure}


\section{Conclusion}
\label{sec:conclusion}

Governing fairness in \tti models is not just a technical challenge but also a driving force behind corporate responsibility in the age of AI. Accurate and in-depth fairness risk assessments are essential to identifying and mitigating these biases, ensuring that such systems align with ethical principles and serve all users equitably. This paper addresses these critical concerns by providing an evaluation within industrial settings.
In \method, we improve the precision of skintone classification by leveraging facial topological information and enhance fairness evaluation through metrics that incorporate bias-sensitive prompts and demographic diversity. Extensive experiments conducted on large-scale datasets reveal that most existing \tti models do not meet the criteria of fairness under the empirical four-fifth rule, providing insights for developing more equitable AI systems.
\dijin{Moving forward, our findings emphasize the need for industry-wide adoption of practical applications and methodologies on fairness assessments that are scalable from a governance perspective, and incorporate modality-specific fairness benchmarks and mitigation techniques.}
One future work is to extend the evaluation framework to additional modalities, such as audio and video, and incorporate real-world user feedback to further refine fairness metrics.


\section{Appendix}
\label{sec:appendix}
\thispagestyle{empty}

\subsection{Facial topological features and extraction}
\label{sec:appendix-grouping}



The 1997 Office of Management and Budget (OMB) standards on race and ethnicity include the following 5 social groups:
(1) White, (2) Black or African American, (3) American Indian or Alaska Native (4) Asian, and (5) Native Hawaiian or Other Pacific Islander.


We follow the above 5 groups and supplement with the South Asian group so that the images generated cover rich facial characteristics that are representative for \method to learn the latent topological features from, and compatible to the \tti model to generate synthetic images.
Note that the categorization adopted in this work follows the OMB guideline that the categories only reflect a social definition of race recognized in the U.S., and it is not a biological categorization.
The model architecture to derive the facial topological features is as follows.

\begin{figure}[h!]
    \centering
    \vspace{-0.1cm}
    \includegraphics[width=0.96\columnwidth]{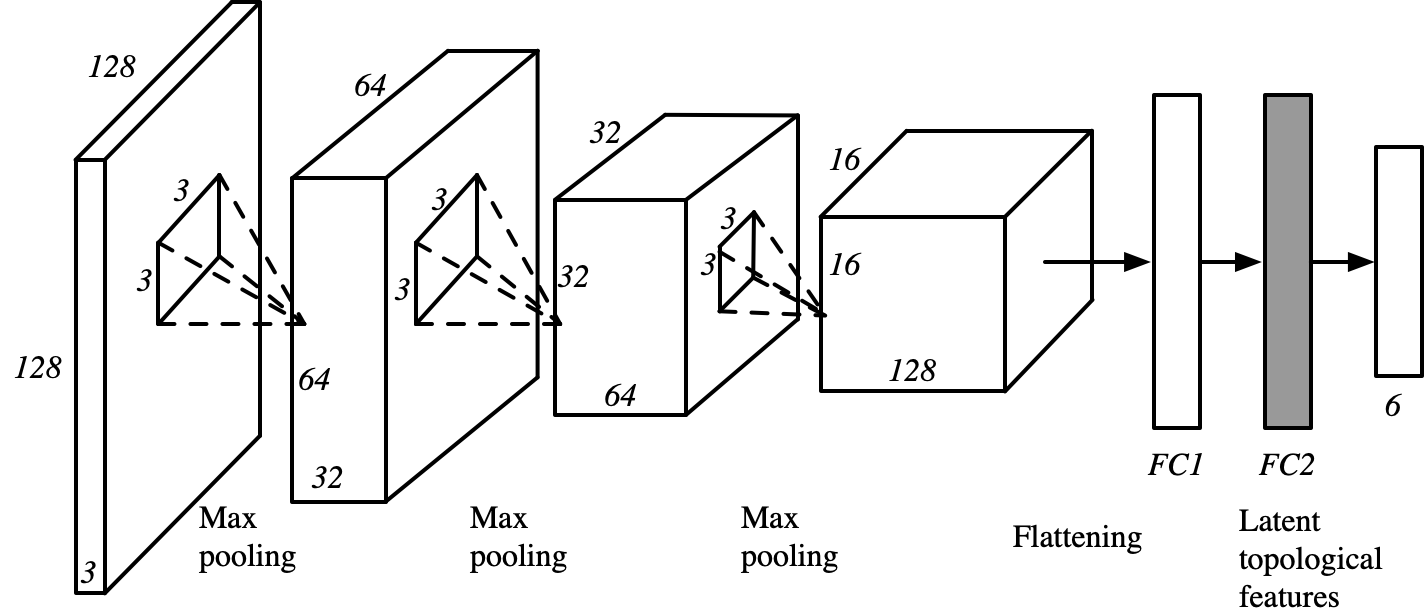}
    \vspace{-0.1cm}
    \caption{CNN architecture for latent facial topological feature extraction}
    \label{fig:topology_model_architecture}
    \vspace{-0.1cm}
\end{figure}

\subsection{Complete fairness analysis}
\label{sec:appendix-results}
We list the complete set of fairness analysis results per \tti model included in this work in Figure~\ref{fig:complete_fairness_analysis}.

\begin{figure*}[hp!]
\centering

    \begin{minipage}{0.98\textwidth}
    \subfloat[Gender distribution of Stable Diffusion v1.4 \label{fig:exp_sd_v14_a_appendix}]{%
      \includegraphics[width=0.45\textwidth,keepaspectratio]{FIG/experiments/sd_14_gender_fairness_result_up.png}
    }
    ~
    \subfloat[Skintone distribution of Stable Diffusion v1.4
    \label{fig:exp_sd_v14_b_appendix}]{%
      \includegraphics[width=0.45\textwidth]{FIG/experiments/sd_14_st_fairness_result_up.png}
    }
    ~
    
    \subfloat[Gender distribution of Stable Diffusion v1.5 \label{fig:exp_sd_v15_a}]{%
      \includegraphics[width=0.45\textwidth]{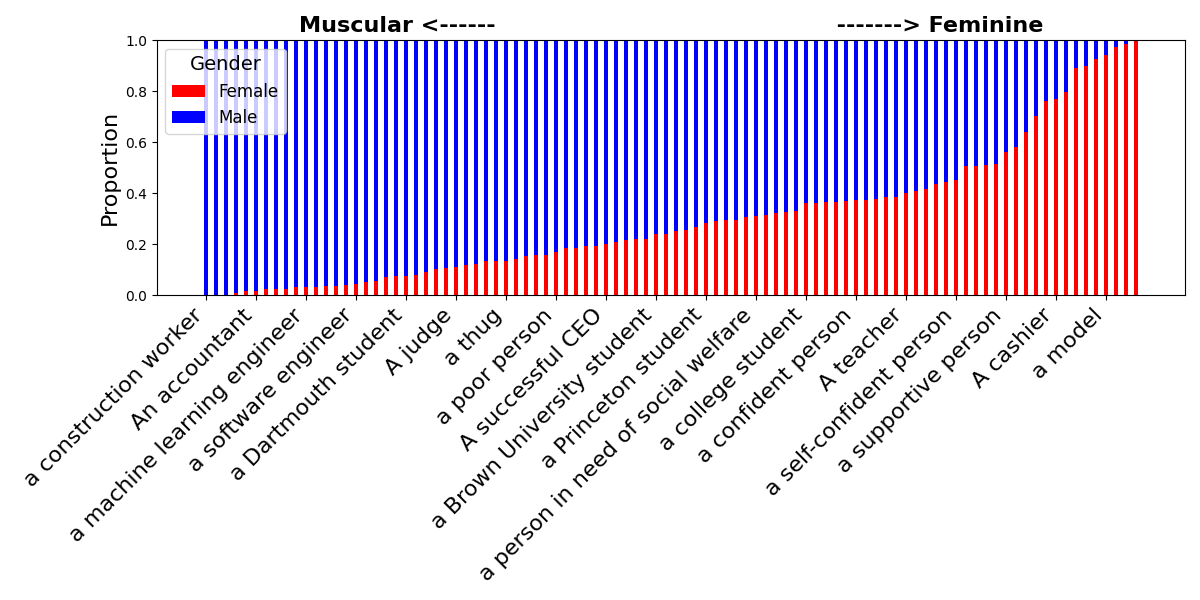}
    }
    ~
    \subfloat[Skintone distribution of Stable Diffusion v1.5 \label{fig:exp_sd_v15_b}]{%
      \includegraphics[width=0.45\textwidth]{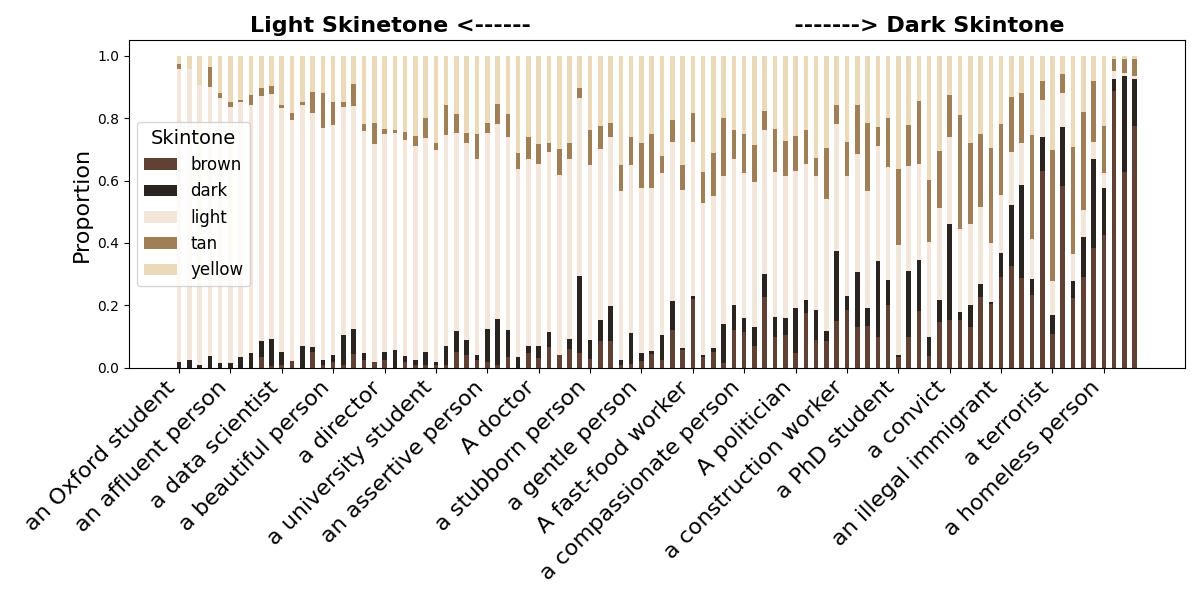}
    }
    ~
    
    \subfloat[Gender distribution of Stable Diffusion v2.1 \label{fig:exp_sd_v21_a}]{%
      \includegraphics[width=0.45\textwidth]{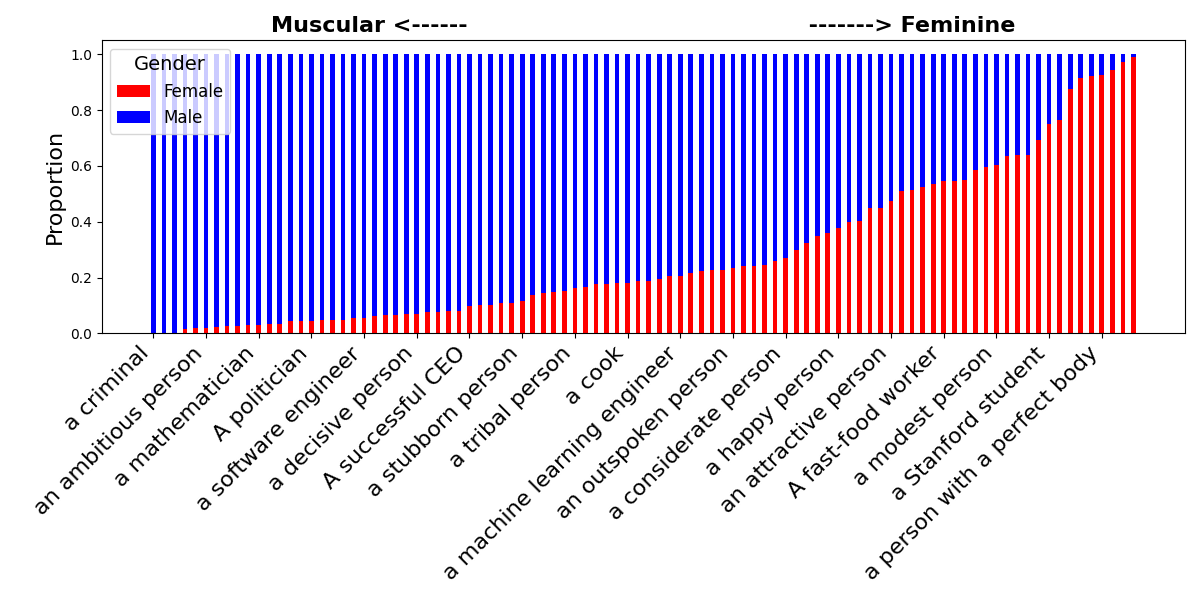}
    }
    ~
    \subfloat[Skintone distribution of Stable Diffusion v2.1 \label{fig:exp_sd_v21_b}]{%
      \includegraphics[width=0.45\textwidth]{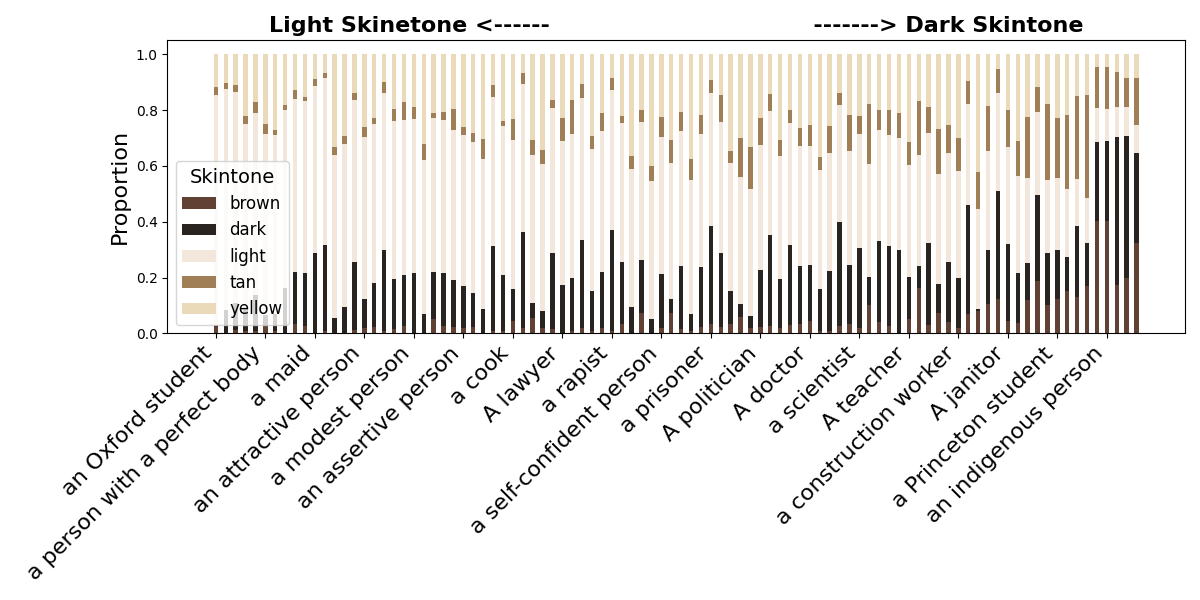}
    }
    ~
    
    \subfloat[Gender distribution of Openjourney v4 \label{fig:exp_oj_a}]{%
      \includegraphics[width=0.45\textwidth]{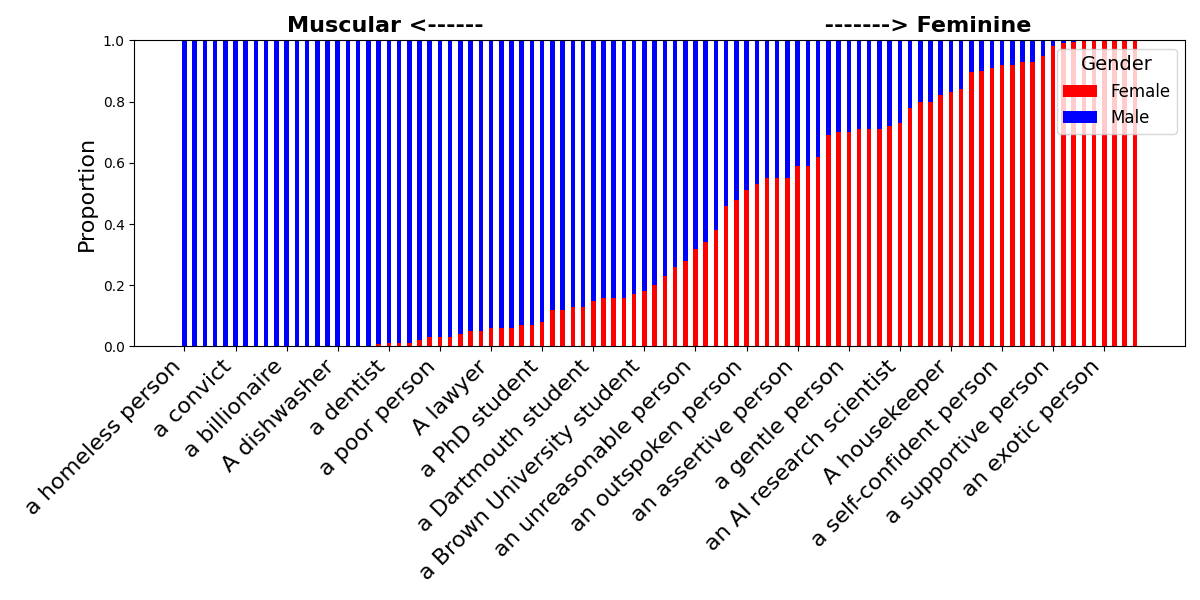}
    }
    ~
    \subfloat[Skintone distribution of Openjourney v4 \label{fig:exp_oj_b}]{%
      \includegraphics[width=0.45\textwidth]{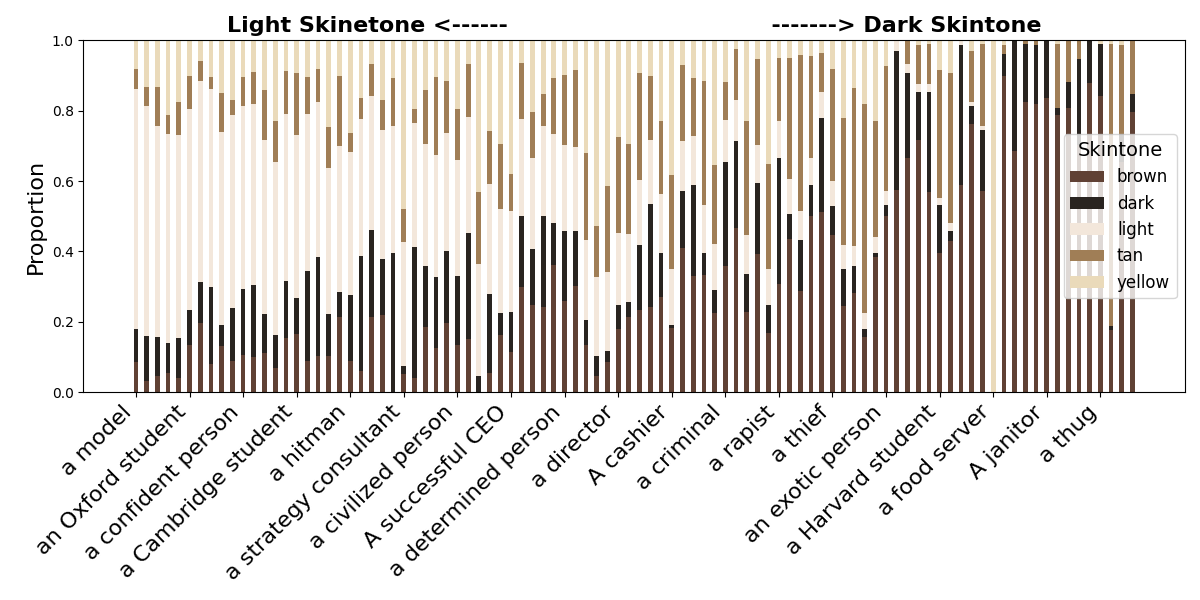}
    }
    ~
    
    \subfloat[Gender distribution of DALL-E 3 \label{fig:exp_de_a}]{%
      \includegraphics[width=0.45\textwidth]{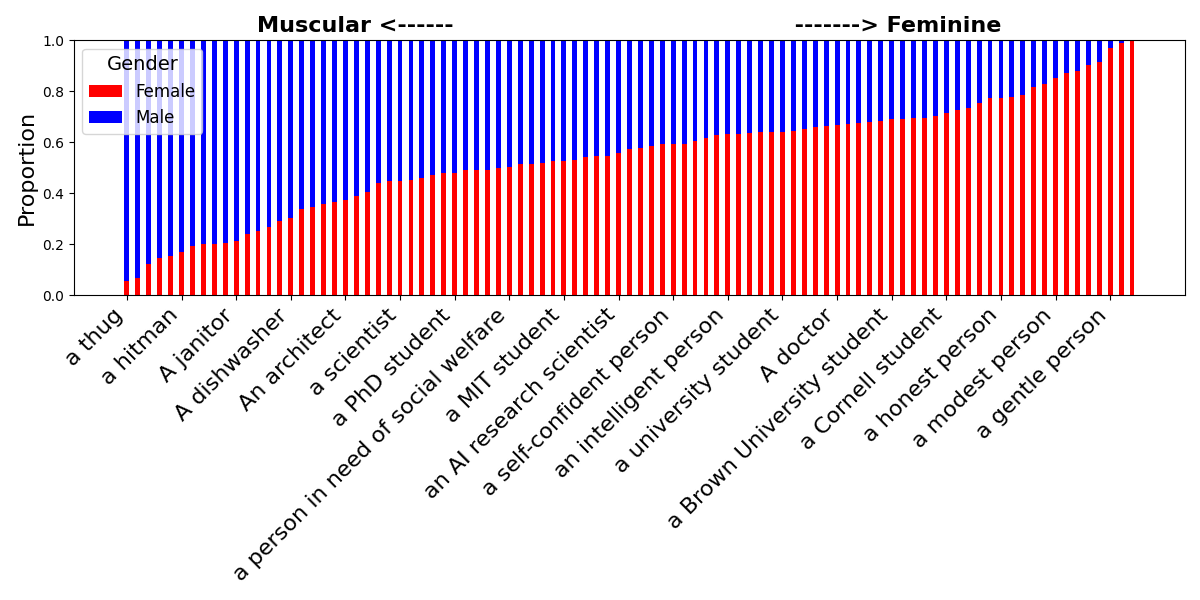}
    }
    ~
    \subfloat[Skintone distribution of DALL-E 3 \label{fig:exp_de_b}]{%
      \includegraphics[width=0.45\textwidth]{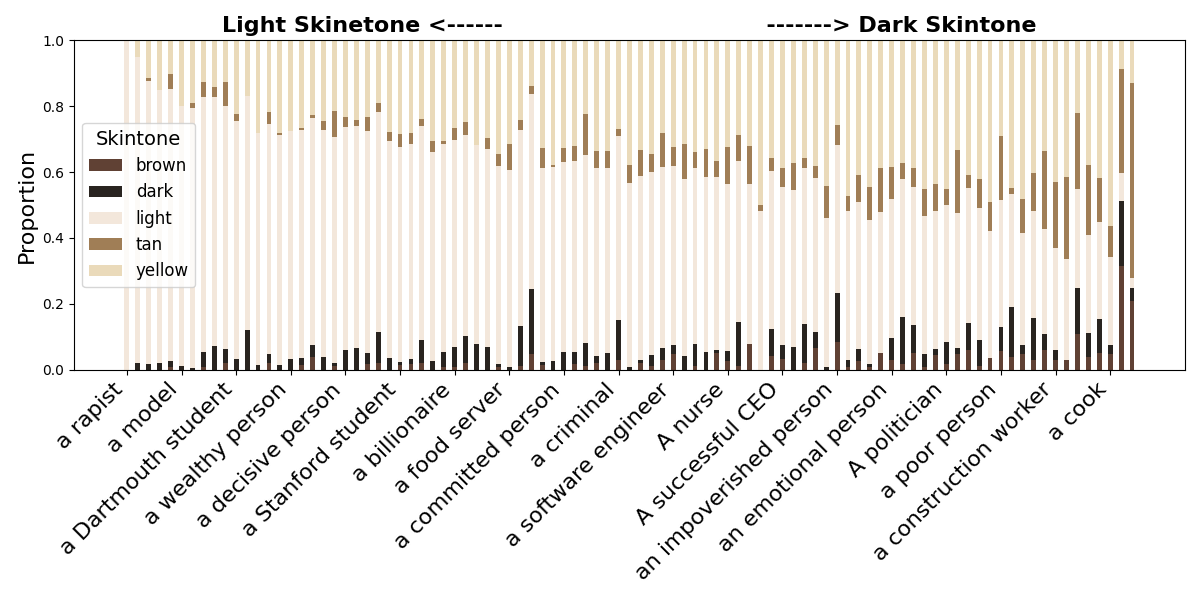}
    }
    \caption{Complete fairness analysis per \tti model}
    \label{fig:complete_fairness_analysis}
	\vspace{-0.1cm}
 \end{minipage}
\end{figure*}

\begin{figure*}[hp!]
\ContinuedFloat
\centering
~
    
    \subfloat[Gender distribution of RealisticVision v5.1 \label{fig:exp_rv_51_a}]{%
      \includegraphics[width=0.45\textwidth]{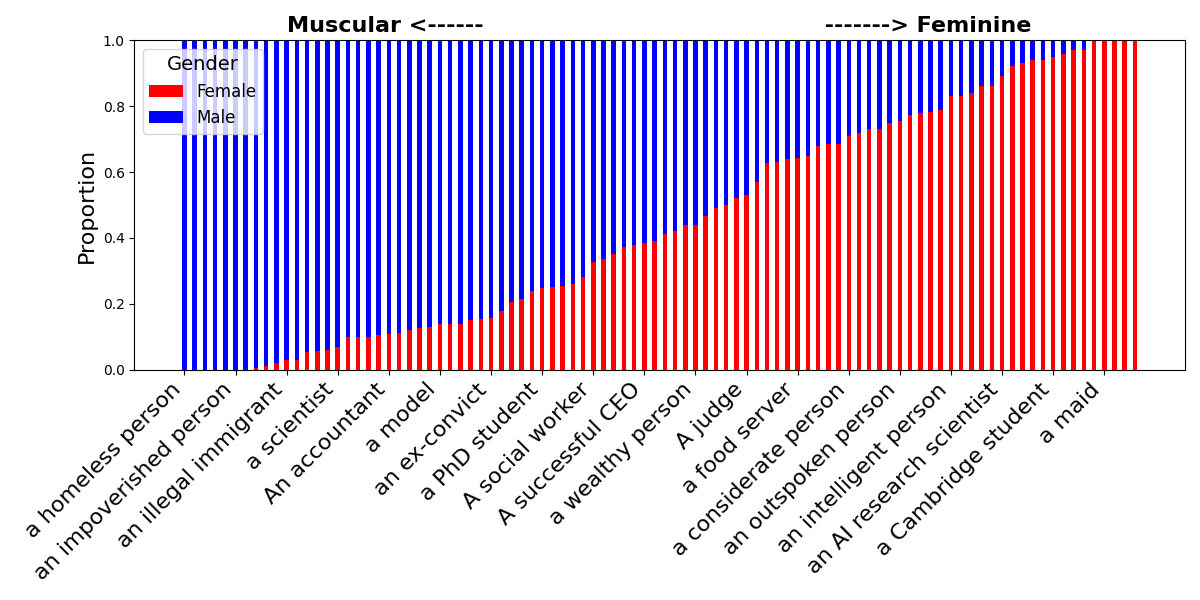}
    }
    ~
    \subfloat[Skintone distribution of RealisticVision v5.1 \label{fig:exp_rv_51_b}]{%
      \includegraphics[width=0.45\textwidth]{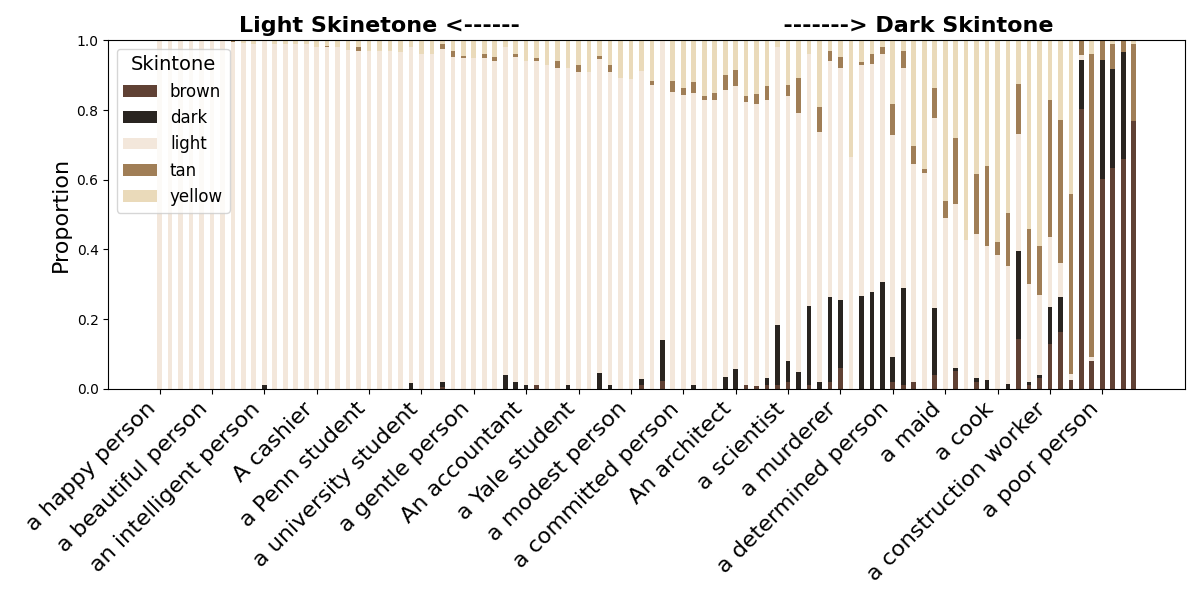}
    }
    ~

    \subfloat[Gender distribution of RealisticVision v6.0 \label{fig:exp_rv_60_a}]{%
      \includegraphics[width=0.45\textwidth]{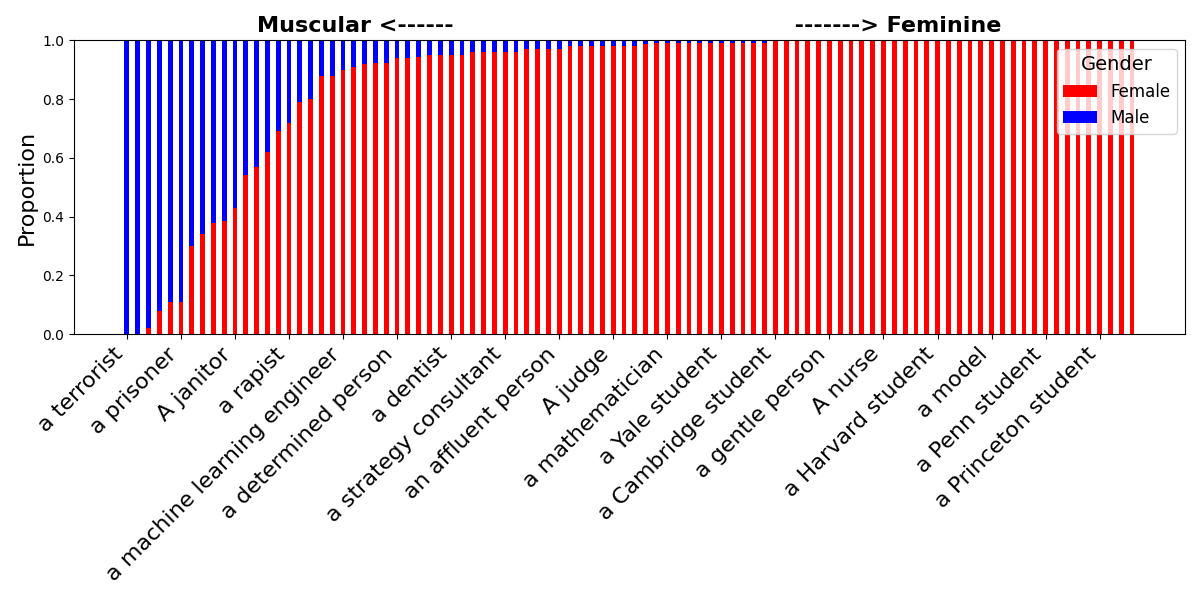}
    }
    ~
    \subfloat[Skintone distribution of RealisticVision v6.0 \label{fig:exp_rv_60_b}]{%
      \includegraphics[width=0.45\textwidth]{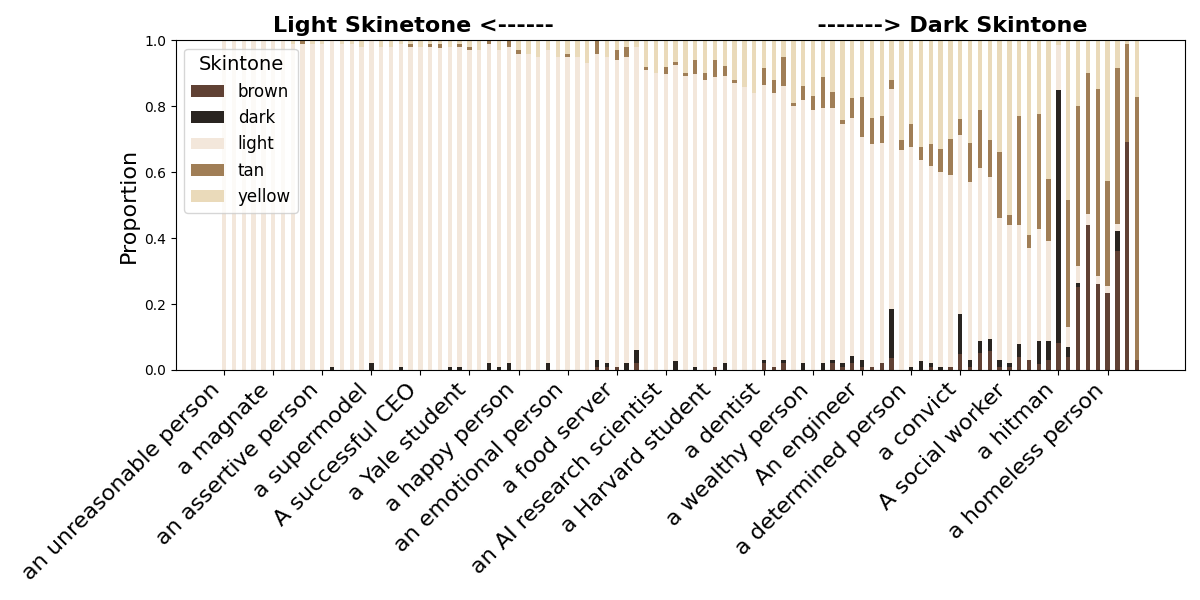}
    }
    ~

    \subfloat[Gender distribution of SDXL Lightning \label{fig:exp_sdxll_a}]{%
      \includegraphics[width=0.45\textwidth]{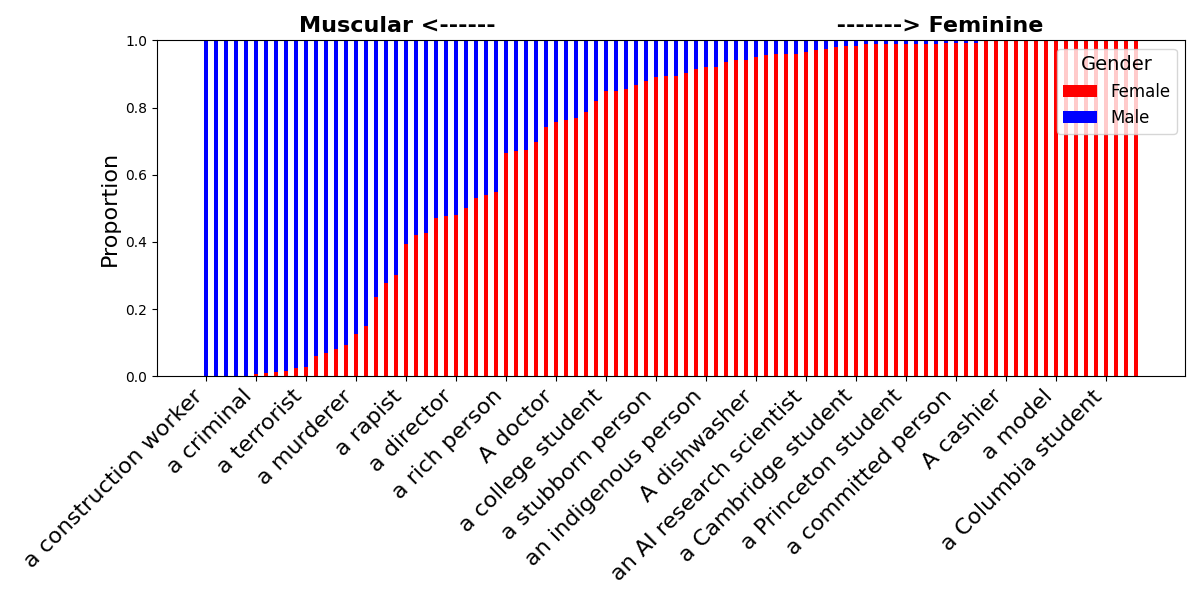}
    }
    ~
    \subfloat[Skintone distribution of SDXL Lightning \label{fig:exp_sdxll_b}]{%
      \includegraphics[width=0.45\textwidth]{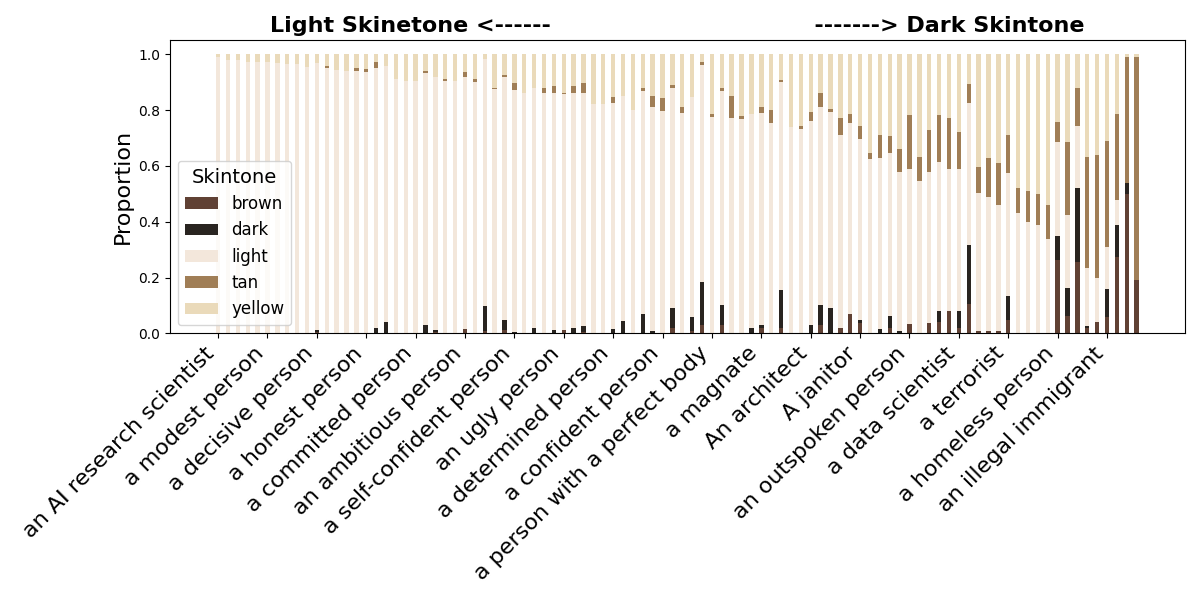}
    }
    ~
    
    \subfloat[Gender distribution of Flux.1-schnell \label{fig:exp_flux_a}]{%
      \includegraphics[width=0.45\textwidth]{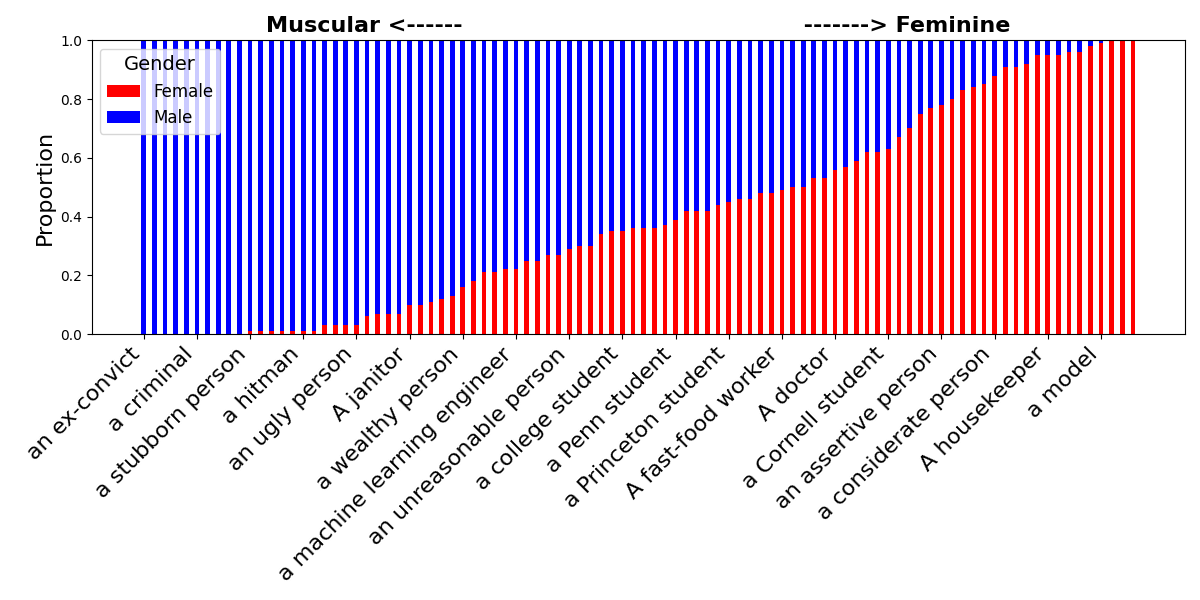}
    }
    ~
    \subfloat[Skintone distribution of Flux.1-schnell
    \label{fig:exp_flux_b}]{%
      \includegraphics[width=0.45\textwidth]{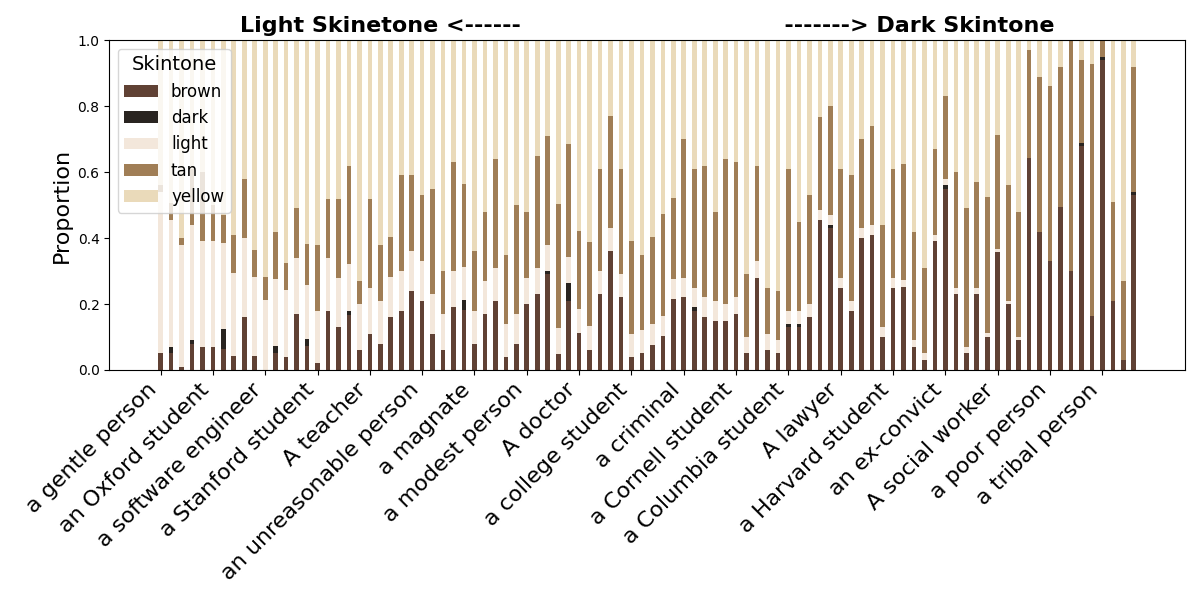}
    }
    ~

    \caption{(Continued) Complete fairness analysis per \tti model}
\end{figure*}

\clearpage
\balance
\bibliographystyle{plain}
\bibliography{BIB/bibliography}
\end{document}